\documentclass{article}

\usepackage[preprint]{neurips_2026}
\usepackage[utf8]{inputenc}
\usepackage[T1]{fontenc}
\usepackage{hyperref}
\usepackage{url}
\usepackage{booktabs}
\usepackage{amsfonts}
\usepackage{nicefrac}
\usepackage{microtype}
\usepackage{xcolor}
\usepackage{amsmath}
\usepackage{amssymb}
\usepackage{graphicx}

\usepackage{tikz}
\usetikzlibrary{arrows,cd,decorations.markings,shapes.geometric,shapes}
\usetikzlibrary{positioning,calc,arrows.meta,fit}
\usetikzlibrary{decorations.pathreplacing}

\newtheorem{definition}{Definition}

\usepackage{pgfplots}
\usepackage{subcaption}
\usetikzlibrary{positioning,decorations.pathreplacing}
 \pgfplotsset{compat=1.18}

\usepackage{algorithm}
\usepackage{algpseudocode}

\newcommand{\softmax}{\operatorname{softmax}}
\newcommand{\detach}{\operatorname{detach}}

\title{Kan Extension Transformers:\\A Categorical Unification of  Attention, Diffusion, and Predict-Detach Self-Conditioning}

\author{%
  Sridhar Mahadevan\thanks{Academic affiliation: Research Professor, University of Massachusetts, Amherst; See webpage at \url{https://people.cs.umass.edu/~mahadeva/Site/About_Me.html}} \\
  Adobe Research\\
  San Jose, CA\\
  \texttt{smahadev@adobe.com} \\
}

\begin{document}

\maketitle

\begin{abstract}
We propose Kan Extension Transformers (KETs) as a categorical design
language for a diverse group of Transformer implementations. A layer can be
viewed generally as a weighted structured extension operator: attention uses
token neighborhoods, geometric mixing uses sparse incidences, and KET uses
simplicial sources. This operator is an actual enriched left Kan extension
only when the source values are functorial, the weights are representable
hom-objects (or the specified profunctor action), and aggregation realizes
the corresponding coend; otherwise ``Kan-style'' denotes an interpretation
rather than an identity theorem. Predict-detach blocks gradients through a
predictive carrier and avoids transporting teacher-forced hidden states, but
detach alone does not make a noncausal update strictly autoregressive: every
carrier consumed at target \(t\) must also be measurable from the prefix
available at \(t\). We evaluate 12 implementations on Penn Treebank,
WikiText-2, and WikiText-103 across strict-causal and self-conditioned
regimes, using widths \(d=64,256\) and depths \(L=2,8,16\) across the
reported studies. Quadratic KET is strongest among the compared
strict-causal architectures on WikiText-2 and WikiText-103; the largest
cross-regime gains arise from additional self-conditioning information, not
neighborhood design alone.
\end{abstract}

\section{Introduction}

A foundational problem in machine learning, already visible in Gold's
formalization of language identification in the limit
\citep{GOLD1967447}, is \emph{generalization}: extending information
from a small or local domain to a larger structured domain. Category
theory reframes this as the extension of a \emph{functor}, a structured
map on both objects and arrows, along a structural map
\citep{maclane:71,riehl2017category,richter2020categories}. Once source
values and target structure are organized functorially, \emph{left} and
\emph{right Kan extensions} provide principled extension operations.

Attention extends token evidence into contextualized representations
\citep{DBLP:conf/nips/VaswaniSPUJGKP17}, while diffusion-style models
refine partial or noisy structure through denoising \citep{ho2020denoising}.
KETs give a common structured-extension language for both: once we choose
source neighborhoods and values on them, a contextual update transports
local objects to token positions. Attention uses singleton neighborhoods,
Geometric Transformer style mixing uses sparse local incidences, and KET
uses simplices and their incidences. The term \emph{Kan extension} is exact
for the representable enriched cases characterized in
Section~\ref{app:kan_extensions}; a generic learned weighted sum need not
satisfy that universal property.

This viewpoint matters most when we ask what values are extended. Ordinary
teacher-forced hidden states can leak gold future information through
noncausal neighborhoods. \emph{Detached predictive carriers} replace those
hidden states by model predictions and block the auxiliary backward path.
This is useful self-conditioning, but its forward causal status still
depends on indexing: a target may consume only carriers constructed from
information available in that target's prefix if the model is to remain
strictly autoregressive.

We focus the paper around three claims.
\begin{enumerate}
\item KET is a weighted left-Kan-style extension over simplicial
neighborhoods; it is a genuine enriched Kan extension in specified
representable cases, while arbitrary learned attention remains an analogy.
\item Predict-detach separates forward carrier reuse from backward credit
assignment. Strict autoregression additionally requires a target-relative
filtration condition; detach by itself is not a proof of causal validity.
\item We compare 12 Transformer variants on PTB, WikiText-2, and
WikiText-103. The studies cover widths \(d=64,256\) and depths
\(L=2,8,16\), although not as one full factorial benchmark. Strict-causal
quadratic KET is strongest on the two larger corpora, while the dominant
cross-regime effect comes from self-conditioning.
\end{enumerate}

\subsection{Background Materials}

This paper builds on six years of previous research on a categorical foundation for Artificial General Intelligence (AGI) -- now available as a $600$-page book complete with a Lean-4 verification of the theoretical results -- which contains a deeper theoretical study of Kan Extension Transformers  \citep{mahadevanCatAGIBook}. There is a chatbot {\tt CLIFF} implemented using a novel deep learning language called \textsc{Functorflow} that provides implementation details of the Kan Extension Transformers (KETs) described in this paper \citep{mahadevanCLIFFCatAgi}. KET models were used in two state of the art deep causal research systems, \textsc{Democritus} \citep{mahadevan2025largecausalmodelslarge} and \textsc{Prometheus} \citep{mahadevan2026prometheusautomatingdeepcausal}.  Supplementary Materials give the categorical details (Section~\ref{app:kan_extensions}), information-regime analysis for GT (Section~\ref{app:info_regimes}), and reproducibility details.

\section{Neighborhood Systems and Kan-Style Aggregation}

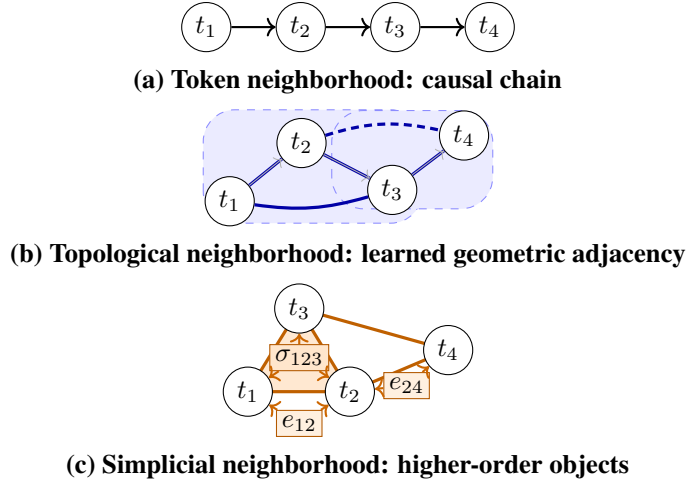
\begin{figure}[t]
\centering

\begin{tikzpicture}[
    token/.style={circle,draw,fill=white,inner sep=1.5pt,minimum size=6.5mm},
    causal/.style={->,thick}
]

\node[token] (t1) at (0,0) {$t_1$};
\node[token] (t2) at (1.25,0) {$t_2$};
\node[token] (t3) at (2.50,0) {$t_3$};
\node[token] (t4) at (3.75,0) {$t_4$};

\draw[causal] (t1) -- (t2);
\draw[causal] (t2) -- (t3);
\draw[causal] (t3) -- (t4);

\end{tikzpicture}

\vspace{0.25em}
\textbf{(a) Token neighborhood: causal chain}

\vspace{0.5em}

\begin{tikzpicture}[
    token/.style={circle,draw,fill=white,inner sep=1.5pt,minimum size=6.5mm},
    topo/.style={blue!65!black,very thick},
    latent/.style={draw=blue!45,fill=blue!8,dashed,rounded corners=12pt}
]

\path[latent] (-0.35,-0.24) rectangle (2.55,1.26);
\path[latent] (1.35,-0.05) rectangle (3.48,1.25);

\node[token] (t1) at (0.00,0.05) {$t_1$};
\node[token] (t2) at (0.95,0.82) {$t_2$};
\node[token] (t3) at (2.15,0.20) {$t_3$};
\node[token] (t4) at (3.10,0.92) {$t_4$};

\draw[topo] (t1) -- (t2);
\draw[topo] (t2) -- (t3);
\draw[topo] (t1) to[bend right=12] (t3);
\draw[topo] (t3) -- (t4);
\draw[topo,densely dashed] (t2) to[bend left=15] (t4);

\draw[->,thin,gray!60] (t1) -- (t2);
\draw[->,thin,gray!60] (t2) -- (t3);
\draw[->,thin,gray!60] (t3) -- (t4);

\end{tikzpicture}

\vspace{0.25em}
\textbf{(b) Topological neighborhood: learned geometric adjacency}

\vspace{0.5em}

\begin{tikzpicture}[
    token/.style={circle,draw,fill=white,inner sep=1.5pt,minimum size=6.5mm},
    edgeobj/.style={rectangle,draw=orange!70!black,fill=orange!12,inner sep=2pt},
    face/.style={fill=orange!18,draw=orange!75!black,very thick},
    inc/.style={->,orange!80!black,thick}
]

\coordinate (p1) at (0,0);
\coordinate (p2) at (1.35,0);
\coordinate (p3) at (0.68,1.10);
\coordinate (p4) at (2.65,0.55);

\draw[face] (p1) -- (p2) -- (p3) -- cycle;
\draw[orange!75!black,very thick] (p2) -- (p4);
\draw[orange!75!black,very thick] (p3) -- (p4);

\node[token] (t1) at (0,0) {$t_1$};
\node[token] (t2) at (1.35,0) {$t_2$};
\node[token] (t3) at (0.68,1.10) {$t_3$};
\node[token] (t4) at (2.65,0.55) {$t_4$};

\node[edgeobj] (e12) at (0.68,-0.42) {$e_{12}$};
\node[edgeobj] (e24) at (2.10,0.08) {$e_{24}$};
\node[edgeobj,fill=orange!20] (s123) at (0.68,0.45) {$\sigma_{123}$};

\draw[inc] (s123) -- (t1);
\draw[inc] (s123) -- (t2);
\draw[inc] (s123) -- (t3);
\draw[inc] (e12) -- (t1);
\draw[inc] (e12) -- (t2);
\draw[inc] (e24) -- (t2);
\draw[inc] (e24) -- (t4);

\end{tikzpicture}

\vspace{0.25em}
\textbf{(c) Simplicial neighborhood: higher-order objects}

\caption{
Three neighborhood systems. (a) Attention uses token-level neighborhoods.
(b) TopoCoend adds learned geometric edges in a latent space. (c) KET
adds higher-order simplices, so aggregation may occur over edges, faces,
and motifs rather than only tokens.
}
\label{fig:neighborhood_systems}
\end{figure}

The model families studied in this paper differ less in the form of
their update rule than in the \emph{structure of the neighborhoods}
over which they aggregate.
Standard attention aggregates over tokens.
TopoCoend aggregates over learned topological neighborhoods.
Kan Extension Transformers aggregate over simplicial objects such as
tokens, edges, and higher-order motifs.
This section gives a common categorical language for these choices.
Figure~\ref{fig:neighborhood_systems} illustrates the distinction:
standard attention follows token-level neighborhoods, TopoCoend learns
geometric adjacency in a latent space, and KET adds higher-order
simplicial source objects such as edges and faces.

\subsection{A Common Setup}

Let
\[
\mathcal{T} = \{0,\dots,S-1\}
\]
denote the target token positions in a sequence window of length \(S\).
A model update at token \(t\) is built from a \emph{source neighborhood
system} \(\mathcal{N}\), whose objects may be tokens, edges, simplices,
or learned geometric neighbors.

We write
\[
X : \mathcal{N} \to \mathbf{Vect}
\]
for the assignment of feature vectors to source objects.
Here \(X(\sigma)\) may be a token state, edge embedding, or higher-order
simplex value, and its contribution to target token \(t\) is determined
by a structural weight \(W(t,\sigma)\) arising from attention scores,
topological kernels, or simplicial incidence.

This yields a common weighted aggregation form
\begin{equation}
h'_t
\;\approx\;
\int^{\sigma \in \mathcal{N}} W(t,\sigma)\otimes X(\sigma)
\;\approx\;
\sum_{\sigma \in \mathcal{N}} w(t,\sigma)\,V(\sigma),
\label{eq:generic_kan_pool}
\end{equation}
where the coend notation emphasizes that the update is a structured
weighted extension rather than an arbitrary sum (see Section~\ref{app:kan_extensions} for more details on coend calculus). 

Equation~\eqref{eq:generic_kan_pool} is the computational form shared by
attention, TopoCoend, and KET.
What changes from one model family to another is the choice of
\(\mathcal{N}\) and the origin of the weights \(W(t,\sigma)\).

\subsection{Token Neighborhoods: Attention}

For standard self-attention, the source neighborhood system contains
only token objects:
\[
\mathcal{N}_{\mathrm{attn}} = \{0,\dots,S-1\}.
\]
Each source object is a token \(s\), and the update at target token \(t\)
takes the familiar form
\[
h'_t = \sum_s w(t,s)\,V_s.
\]

In categorical language, this is a weighted left-Kan-style extension in
which source and target are both token positions; KET and TopoCoend
generalize it by enriching the source neighborhood system.

\subsection{Topological Neighborhoods: TopoCoend}

TopoCoend keeps the source objects at the token level but changes the
neighborhood relation.
Instead of using only positional adjacency or dense token-to-token
attention, it learns a low-dimensional geometric representation
\[
z_t = \pi(v_t),
\]
where \(v_t\) is either a hidden state or a detached predictive carrier.
A fuzzy \(k\)-nearest-neighbor graph is then constructed in the latent
space of the \(z_t\)'s.

The resulting \(\mathcal{N}_{\mathrm{topo}}\) is a learned geometric
graph whose weights \(W_{\mathrm{topo}}(t,s)\) are induced by topological
proximity rather than sequence position alone.
The corresponding update is
\begin{equation}
h'_t
=
h_t + \sum_s w_{\mathrm{topo}}(t,s)\,V_s.
\label{eq:topocoend_update}
\end{equation}

TopoCoend is thus weighted aggregation over a learned topological
neighborhood category.

\subsection{Simplicial Neighborhoods: KET}

KET changes not just the weights but the source objects themselves:
\begin{itemize}
    \item \(0\)-simplices: tokens \(t\),
    \item \(1\)-simplices: edges such as \((t-1,t)\),
    \item optional higher simplices: faces, motifs, or larger spans.
\end{itemize}

Let \(\mathcal{N}_{\mathrm{simp}}\) denote this simplicial indexing
category. Each simplex \(\sigma\) has value \(X(\sigma)\), and its
contribution to token \(t\) is controlled by \(W_{\mathrm{simp}}(t,\sigma)\).
The generic KET update is therefore
\begin{equation}
h'_t
=
h_t + \sum_{\sigma \in \mathcal{N}_{\mathrm{simp}}}
w_{\mathrm{simp}}(t,\sigma)\,V(\sigma).
\label{eq:simplicial_update}
\end{equation}

Thus KET aggregates over \emph{higher-order source objects}, not just
individual tokens, generalizing both attention and local geometric mixing.

\subsection{Quadratic and Incidence-Restricted KET}

The simplicial update in Eq.~\eqref{eq:simplicial_update} admits two
important realizations.

\paragraph{Quadratic KET.}
In the global or quadratic variant, every token may aggregate from
every simplex:
\begin{equation}
h'_t
=
h_t +
\sum_{\sigma \in \mathcal{N}_{\mathrm{simp}}}
w(t,\sigma)\,V(\sigma),
\qquad
w(t,\sigma)=\operatorname{softmax}(Q_t^\top K_\sigma).
\label{eq:quadratic_ket}
\end{equation}
This is a second attention-like kernel, now defined over simplices
rather than only over tokens.
Its complexity is typically \(O(S^2)\) in sequence length.

\paragraph{Incidence-restricted KET.}
In the incidence-restricted variant, only simplices incident to a token
may contribute:
\begin{equation}
h'_t
\;\approx\;
h_t + \sum_{\sigma \ni t} \phi\bigl(V(\sigma)\bigr).
\label{eq:incidence_ket}
\end{equation}
For edge-only models this becomes
\begin{align}
e_t &= \psi([v_{t-1},v_t]), \\
h'_t &= h_t + \phi(e_t)
\qquad \text{(causal incidence)}.
\end{align}
This reduces the complexity to \(O(S)\) and makes the connection to Geometric Transformers (GT) more 
transparent.

\subsection{Relationship to Geometric Transformers}

The geometric branch of a Geometric Transformer is an
incidence-restricted Kan-style update in which the message map
is implemented by a local convolutional or message-passing operator.
This yields the following hierarchy:

\begin{itemize}
    \item \textbf{Attention:} weighted extension over tokens.
    \item \textbf{TopoCoend:} weighted extension over learned topological neighborhoods.
    \item \textbf{Incidence-restricted KET:} weighted extension over incident simplices.
    \item \textbf{Quadratic KET:} weighted extension over all simplices with a learned global kernel.
    \item \textbf{GT:} an efficient incidence-restricted special case of KET.
\end{itemize}

This hierarchy explains the later expressive and runtime tradeoffs. It
also clarifies our limited diffusion claim: predictive or denoising
carriers turn the same transport rule into structured self-conditioning
over partially specified future content.

\section{Predict-Detach and the Causal Filtration}

\subsection{Three Information Regimes}

The experiments compare three regimes (for a more detailed discussion of information regimes, please see Section~\ref{app:info_regimes} in the Supplementary Materials). 

\paragraph{Strict-causal.}
Both neighborhoods and values are restricted to prefix-valid information. This is the standard autoregressive setting.

\paragraph{Gold noncausal.}
The model is allowed to mix teacher-forced future hidden states through noncausal neighborhoods. This regime is invalid for language modeling but is useful as a leakage diagnostic.

\paragraph{Predict-detach.}
The model transports detached predictions rather than teacher-forced hidden
states. We distinguish a \emph{prefix-valid} version, in which every carrier
used at target \(t\) is computed from information available by \(t\), from a
\emph{noncausal self-conditioning} version that mixes carriers generated at
future-indexed positions. Only the former is strictly autoregressive.

\subsection{Predictive Carriers}

For token \(t\), let \(\ell_t\) be the next-token logits produced from its causal hidden state. We build a predictive carrier
\begin{equation}
\hat e_t
=
\detach\!\left(\softmax(\ell_t / T)\, E\right),
\label{eq:predict_detach}
\end{equation}
where \(E\) is the embedding matrix and \(T\) is a temperature. The KET
layer may then transport \(\hat e_t\) rather than teacher-forced hidden
states. Future positions can therefore exchange \emph{predicted}
content through simplicial neighborhoods even though they never
exchange teacher-forced future hidden states.

\subsection{Target-Relative Filtration Criterion}

Let \(\mathcal F_t=\sigma(x_{\leq t})\) be the information available when
predicting \(x_{t+1}\). A carrier generated at source position \(s\),
\[
\hat e_s
=
\detach\!\left(\softmax(\ell_s/T)E\right),
\]
is generally \(\mathcal F_s\)-measurable. An update used to predict at
position \(t\) is strictly autoregressive only if every transported carrier
is \(\mathcal F_t\)-measurable. This holds, for example, for carriers with
\(s\leq t\), including the shifted-previous construction, or when a carrier
is recomputed solely from the target prefix. If \(s>t\), then
\(\hat e_s\) may depend on gold tokens \(x_{t+1:s}\) even though it is
detached. Such a run is self-conditioned and avoids direct hidden-state
transport, but it is not a strict autoregressive likelihood evaluation.

\subsection{Why Detach Matters}

There are two distinct reasons for the detach in \eqref{eq:predict_detach}.

\paragraph{No direct teacher-forced hidden-state carrier.}
At its own source index, the carrier is a prediction rather than the
teacher-forced hidden state itself. This removes one direct leakage channel,
but it does not establish target-relative prefix validity when \(s>t\).

\paragraph{No leakage gradient.}
Detaching blocks the auxiliary branch from becoming a backdoor through which the model could cheaply encode targets into the very carriers later consumed by noncausal aggregation.

Predict-detach is therefore an optimization or modal boundary, not by itself
a causal boundary. Operationally, it resembles internal prompt repetition
or one-step denoising: the model writes down a guess and reuses it as
structured context, while the auxiliary loss cannot backpropagate through
that guess. Forward causal validity is checked separately by the filtration
criterion above.

\section{Experiments comparing Information Regimes}

\subsection{Setup}

We evaluate KET on Penn Treebank, WikiText-2, and WikiText-103. The
headline next-token comparisons use context length \(128\), \(L=2\), and
\(d=256\); a matched width check uses \(d=64\), and the
structured-completion study uses \(L=8\) and \(L=16\) at \(d=64\).
Thus the paper covers depths \(2,8,16\) and widths \(64,256\), but it
does not claim a full depth-by-width factorial scaling study. Headline
tables compare a causal Transformer, GT, quadratic KET, and
incidence-restricted KET; Figure~\ref{fig:wt103} overlays TopoCoend
trajectories. Self-conditioned runs use predictive carriers, and
gold-noncausal models are retained only as leakage diagnostics. The
supplementary repository contains harnesses, configs, summaries, logs,
and checkpoints.

\begin{table}[t]
\centering
\small
\caption{Depth and width coverage. The deeper runs evaluate structured
completion rather than the headline next-token objective, so the three rows
are complementary rather than a single factorial sweep.}
\label{tab:scale_coverage}
\begin{tabular}{lccc}
\toprule
Study & Depth \(L\) & Width \(d\) & Objective \\
\midrule
Headline LM comparison & 2 & 256 & next-token \\
Width/capacity check & 2 & 64, 256 & next-token \\
Higher-depth comparison & 8, 16 & 64 & block/direct and denoise \\
\bottomrule
\end{tabular}
\end{table}

\subsection{Strict-Causal Results}

\begin{figure*}[p]
\centering
\begin{minipage}{0.7\textwidth}
\centering
\includegraphics[width=\linewidth]{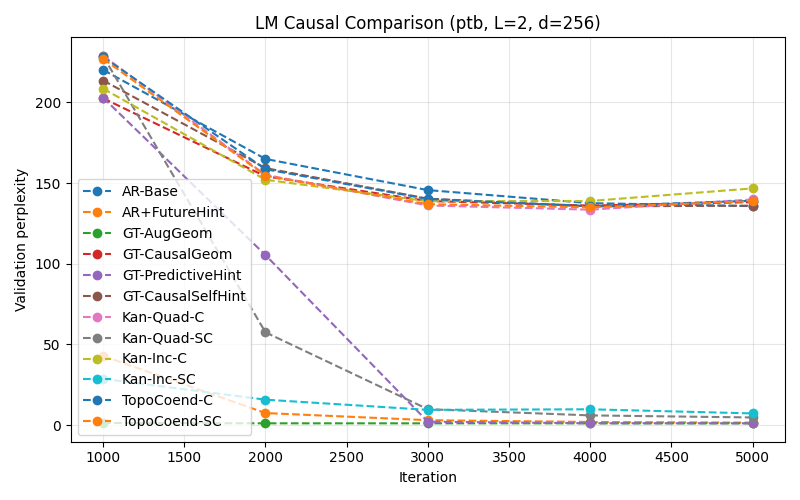}

\small PTB
\end{minipage}
\hfill
\begin{minipage}{0.7\textwidth}
\centering
\includegraphics[width=\linewidth]{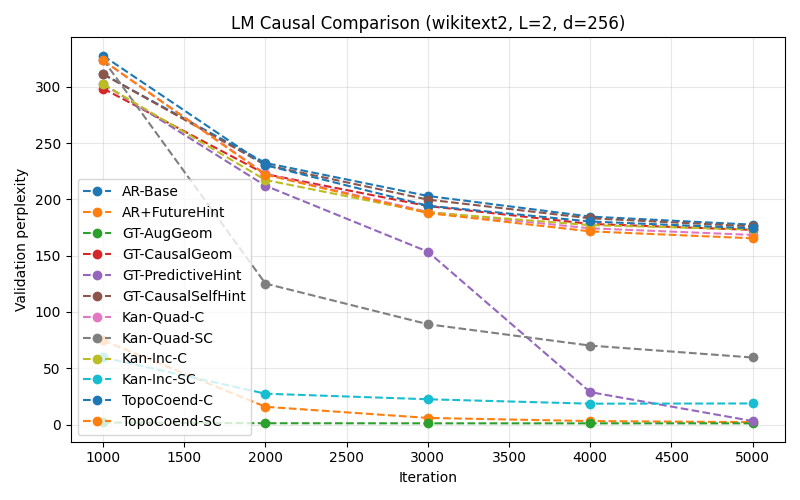}

\small WikiText-2
\end{minipage}
\hfill
\begin{minipage}{0.7\textwidth}
\centering
\includegraphics[width=\linewidth]{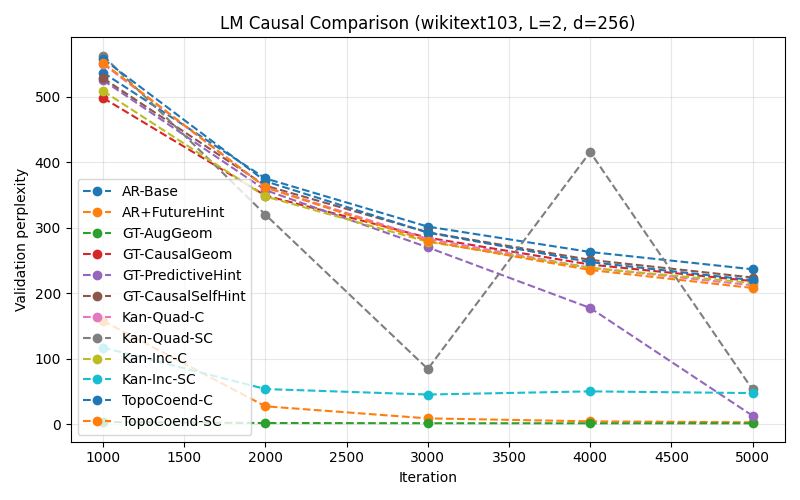}

\small WikiText-103
\end{minipage}
\caption{Validation perplexity across datasets. The headline tables quantify Transformer, GT, and KET comparisons; the plots additionally show TopoCoend for qualitative reference. On PTB, the Transformer baseline is strongest among the tabulated strict-causal models. On WikiText-2 and WikiText-103, quadratic KET is the strongest tabulated strict-causal model. Self-conditioned GT/KET variants improve sharply, TopoCoend changes more modestly, and explicit future-information leakage diagnostics collapse.}
\label{fig:wt103}
\end{figure*}

Table \ref{tab:strict} shows the strict-causal comparison. On PTB the
plain Transformer remains strongest, while on WikiText-2 and
WikiText-103 quadratic KET is best among the tabulated causal
architectures. The TopoCoend curve in Figure~\ref{fig:wt103} remains in
the same broad causal band. Incidence-restricted KET is slightly weaker
but close on the larger datasets, consistent with an efficient sparse
approximation to the richer quadratic neighborhood system.

\begin{table}[t]
\centering
\caption{Strict-causal test perplexity. Lower is better.}
\label{tab:strict}
\begin{tabular}{lccc}
\toprule
Model & PTB & WT2 & WT103 \\
\midrule
Transformer & \textbf{124.47} & 163.92 & 232.52 \\
GT-Causal & 127.17 & 157.74 & 215.69 \\
KET-Quad-C & 133.37 & \textbf{156.42} & \textbf{210.30} \\
KET-Inc-C & 137.19 & 161.12 & 213.76 \\
\bottomrule
\end{tabular}
\end{table}

\subsection{Predict-Detach Self-Conditioning Results}

\begin{table}[t]
\centering
\caption{Predict-detach self-conditioning ablation. These results form a
separate information regime and are not strict-autoregressive likelihood
comparisons unless the target-relative filtration condition is enforced.}
\label{tab:pd}
\begin{tabular}{lccc}
\toprule
Model & PTB & WT2 & WT103 \\
\midrule
KET-Quad-C & 133.37 & 156.42 & 210.30 \\
KET-Quad-PD & 31.43 & 38.23 & 51.89 \\
KET-Inc-C & 137.19 & 161.12 & 213.76 \\
KET-Inc-PD & \textbf{6.54} & \textbf{19.08} & \textbf{47.17} \\
\midrule
GT-PD & 1.05 & 1.59 & 12.84 \\
\bottomrule
\end{tabular}
\end{table}

Table \ref{tab:pd} isolates the regime effect. Replacing hidden-state
carriers by detached predictive carriers improves every KET architecture:
quadratic KET drops from \(133.37\) to \(31.43\) on PTB, \(156.42\) to
\(38.23\) on WT2, and \(210.30\) to \(51.89\) on WT103; the
incidence-restricted variant improves even more on PTB and WT2. The
strongest self-conditioned baseline is GT-PD, while TopoCoend improves
only modestly. Thus the largest numerical gain comes from the information
regime itself; the neighborhood family determines how that gain is
expressed. Because the noncausal variants may consume future-indexed
carriers, these numbers demonstrate self-conditioned completion rather than
strict autoregressive language-model improvement.

\subsection{Leakage Diagnostic}

When noncausal neighborhoods receive teacher-forced future hidden states,
test perplexity collapses toward \(1\). These leakage diagnostics show that
causal validity is determined not by the graph or by detach alone, but by
the target-relative information used to construct every transported value.

\section{Experiments on Block Denoising as Structured Completion}

A first generalization of standard autoregressive prediction is to
predict not a single next token, but an entire future block at once.
Instead of producing only \(x_{t+1}\), the model predicts $x_{t+1:t+B}$, 
where \(B\) is the block size. The corresponding prediction map is
\begin{equation}
f : C_t \longrightarrow \Sigma^B,
\label{eq:block_product_map}
\end{equation}
where \(C_t\) is the causal context at position \(t\) and
\(\Sigma^B\) denotes the \(B\)-fold product of the vocabulary object:
\[
\Sigma^B
=
\Sigma \times \Sigma \times \cdots \times \Sigma .
\]

Categorically, direct block prediction therefore replaces the usual
single-token target \(\Sigma\) by a structured \emph{product object}.
The model must map the context \(C_t\) into a tuple of future tokens in
one step rather than constructing that future sequentially.  Direct block prediction is challenging: a different approach is to change
not the architecture, but the \emph{information regime}.
Instead of predicting the future block from context alone, we provide
the model with a corrupted or partially specified version of that block
and ask it to reconstruct the original:
\begin{equation}
f : (C_t,\tilde{x}_{t+1:t+B}) \longrightarrow x_{t+1:t+B},
\end{equation}
where \(\tilde{x}_{t+1:t+B}\) is a noisy, masked, or otherwise corrupted
version of the future block. The corrupted block already carries partial structure, and the model
learns to extend that partial object to a full consistent one.
This suggests a
complementary viewpoint to left-Kan aggregation:
where left Kan extensions capture the aggregation of evidence from local
sources, denoising naturally points toward a \emph{right-Kan intuition},
in which the task is to find a value or structure that is compatible
with multiple local constraints at once. 

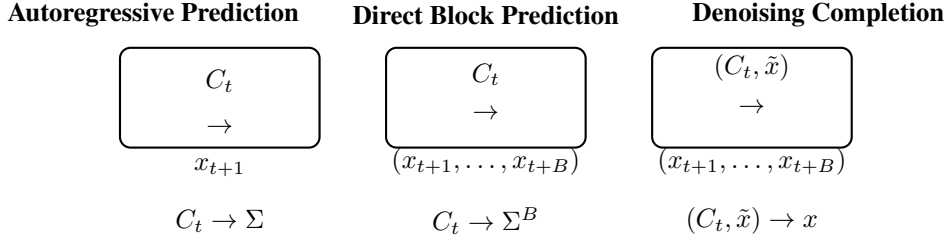
\begin{figure}[t]
\centering
\begin{tikzpicture}[scale=0.88]

\node at (-5,2.5) {\textbf{Autoregressive Prediction}};
\draw[rounded corners, thick] (-5.5,0.5) rectangle (-2.5,2);
\node at (-4,1.5) {$C_t$};
\node at (-4,0.8) {$\rightarrow$};
\node at (-4,0.2) {$x_{t+1}$};
\node at (-4,-0.6) {$C_t \rightarrow \Sigma$};

\node at (0,2.5) {\textbf{Direct Block Prediction}};
\draw[rounded corners, thick] (-1.5,0.5) rectangle (1.5,2);
\node at (0,1.6) {$C_t$};
\node at (0,1.0) {$\rightarrow$};
\node at (0,0.3) {$(x_{t+1},\dots,x_{t+B})$};
\node at (0,-0.6) {$C_t \rightarrow \Sigma^B$};

\node at (5,2.5) {\textbf{Denoising Completion}};
\draw[rounded corners, thick] (2.5,0.5) rectangle (5.5,2);
\node at (4,1.7) {$(C_t,\tilde{x})$};
\node at (4,1.1) {$\rightarrow$};
\node at (4,0.3) {$(x_{t+1},\dots,x_{t+B})$};
\node at (4,-0.6) {$(C_t,\tilde{x}) \rightarrow x$};

\end{tikzpicture}
\caption{Three information regimes for language modeling. Direct block
prediction maps context into the product object \(\Sigma^B\); denoising
completion instead fills a partially specified future block.}
\label{fig:structured_lm_regimes}
\end{figure}

\begin{figure}[t]
\centering
\begin{tikzpicture}[scale=0.82]

\node at (-3.5,2.6) {\textbf{Corrupted Block as a Horn}};
\coordinate (A) at (-4.8,0.4);
\coordinate (B) at (-3.4,1.8);
\coordinate (C) at (-2.0,0.4);
\draw[thick] (A) -- (B);
\draw[thick] (B) -- (C);
\draw[dashed, thick] (A) -- (C);
\filldraw[black] (A) circle (2pt) node[below] {$x_{t+1}$};
\filldraw[black] (B) circle (2pt) node[above] {$x_{t+2}$};
\filldraw[black] (C) circle (2pt) node[below] {$x_{t+3}$};
\node at (-3.4,-0.5) {partial structure $\tilde{x}$};
\node at (-3.4,-1.0) {known faces, missing completion};

\draw[->, very thick] (-1.2,1.0) -- (0.6,1.0);
\node at (-0.3,1.35) {denoising / horn filling};

\node at (3.2,2.6) {\textbf{Completed Simplex}};
\coordinate (D) at (1.8,0.4);
\coordinate (E) at (3.2,1.8);
\coordinate (F) at (4.6,0.4);
\draw[thick] (D) -- (E);
\draw[thick] (E) -- (F);
\draw[thick] (D) -- (F);
\fill[gray!15] (D) -- (E) -- (F) -- cycle;
\filldraw[black] (D) circle (2pt) node[below] {$x_{t+1}$};
\filldraw[black] (E) circle (2pt) node[above] {$x_{t+2}$};
\filldraw[black] (F) circle (2pt) node[below] {$x_{t+3}$};
\node at (3.2,-0.5) {completed structure $x$};
\node at (3.2,-1.0) {full simplex / coherent block};

\end{tikzpicture}
\caption{A simplicial view of denoising language modeling. A corrupted
block is a partial simplex, or horn; denoising fills the missing
relations to produce a coherent completion.}
\label{fig:horn_completion_lm}
\end{figure}
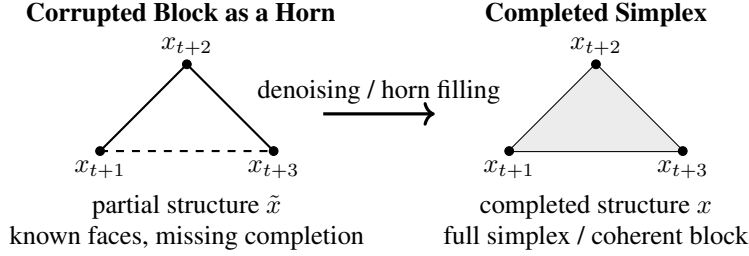

We now report on results with block denoising, a diffusion-style completion task in which
the model predicts a length-4 future block directly or reconstructs it
from a corrupted version. Figures~\ref{fig:structured_lm_regimes} and
\ref{fig:horn_completion_lm} frame the comparison in
Table~\ref{tab:block_main}: direct block prediction is a one-shot map
into \(\Sigma^4\), whereas denoising is a structured completion problem.  This completion process admits a natural simplicial analogy.
A block of tokens may be viewed as a simplex whose vertices correspond
to token positions and whose internal relations encode coherence across
the block.
A corrupted block \(\tilde{x}\) then behaves like a partially specified
simplex: some vertices or relations are present, while others are
missing or uncertain. In simplicial homotopy theory \citep{may1992simplicial}, such partial simplices are often
described as \emph{horns}.
To fill a horn is to extend a partial simplex to a full simplex.
A simplicial set is called a Kan complex when every horn admits such a
filler.

\subsection{Setup}

We ran a structured-completion harness with matched causal Transformer
and incidence-KET backbones on PTB, WikiText-2, and WikiText-103. To
test whether the block objectives change with depth, we report
higher-depth sweeps at \(L=8\) and \(L=16\) with \(d=64\). All block
experiments use context length \(T=128\) and block size \(B=4\). The
direct objective learns $ C_t \longrightarrow \Sigma^4$,
while the denoising objective learns $
(C_t,\tilde{x}_{t+1:t+4}) \longrightarrow x_{t+1:t+4}.
$. 
\noindent For denoising, corruption is applied independently to positions
in the future block using an 8-step schedule
\[
p(s) = 0.05 + 0.45 \cdot \frac{s-1}{7},
\qquad s \in \{1,\dots,8\},
\]
so the corruption rate ranges from \(0.05\) to \(0.50\). We report
first-token perplexity and block perplexity averaged over all four
target offsets.

\subsection{Main Results}

Table \ref{tab:block_main} shows two patterns. Denoising is dramatically
easier than direct block prediction across datasets and depths. Greater
depth mainly helps the hard direct objective: from \(L=8\) to \(L=16\),
KET first-token PPL drops from \(152.25\) to \(143.14\) on PTB, \(202.89\)
to \(174.55\) on WT2, and \(574.06\) to \(472.29\) on WT103. Denoising is
already low-perplexity, so depth and backbone effects are modest.

\begin{table*}[t]
\centering
\small
\caption{Structured language modeling with direct block prediction and block denoising (\(B=4\), context \(=128\), \(d=64\)) on PTB, WikiText-2, and WikiText-103.}
\label{tab:block_main}
\begin{tabular}{llrcc}
\toprule
Dataset & Model & \(L\) & Test first-PPL & Test block-PPL \\
\midrule
PTB & TF-Block-4 & 8 & 155.95 & 332.38 \\
PTB & KET-Block-4 & 8 & 152.25 & 327.62 \\
PTB & TF-Denoise-4 & 8 & 4.21 & 4.56 \\
PTB & KET-Denoise-4 & 8 & 4.21 & 4.55 \\
\midrule
PTB & TF-Block-4 & 16 & 144.14 & 316.65 \\
PTB & KET-Block-4 & 16 & 143.14 & 315.25 \\
PTB & TF-Denoise-4 & 16 & 4.17 & 4.54 \\
PTB & KET-Denoise-4 & 16 & 4.18 & 4.55 \\
\midrule
WikiText-2 & TF-Block-4 & 8 & 194.11 & 362.19 \\
WikiText-2 & KET-Block-4 & 8 & 202.89 & 369.77 \\
WikiText-2 & TF-Denoise-4 & 8 & 4.42 & 4.80 \\
WikiText-2 & KET-Denoise-4 & 8 & 4.44 & 4.81 \\
\midrule
WikiText-2 & TF-Block-4 & 16 & 177.87 & 343.04 \\
WikiText-2 & KET-Block-4 & 16 & 174.55 & 341.37 \\
WikiText-2 & TF-Denoise-4 & 16 & 4.35 & 4.73 \\
WikiText-2 & KET-Denoise-4 & 16 & 4.35 & 4.72 \\
\midrule
WikiText-103 & TF-Block-4 & 8 & 539.92 & 839.24 \\
WikiText-103 & KET-Block-4 & 8 & 574.06 & 859.31 \\
WikiText-103 & TF-Denoise-4 & 8 & 5.83 & 6.07 \\
WikiText-103 & KET-Denoise-4 & 8 & 5.85 & 6.09 \\
\midrule
WikiText-103 & TF-Block-4 & 16 & 491.12 & 796.10 \\
WikiText-103 & KET-Block-4 & 16 & 472.29 & 782.34 \\
WikiText-103 & TF-Denoise-4 & 16 & 5.69 & 5.96 \\
WikiText-103 & KET-Denoise-4 & 16 & 5.74 & 6.01 \\
\bottomrule
\end{tabular}
\end{table*}

These results echo predict-detach. Direct block prediction asks the model
to select an element of \(\Sigma^4\) from context alone; denoising supplies
a partial future object and asks for completion. That information-regime
change dominates the backbone choice, reducing first-token perplexity by
one to two orders of magnitude.

\section{Related Work}

We summarize related work along three lines. Diffusion and denoising
Transformers pair refinement objectives with Transformer architectures
\citep{ho2020denoising,peebles2023scalable,li2022diffusionlm}; our
connection is conceptual rather than score-based. Higher-order,
hypergraph, and topological attention models enrich the aggregation
domain
\citep{hajij2022topological,goh2022simplicial,battiloro2024gsan,zhang2022hegel},
and sheaf-based neural models show that algebraic-topological structure
changes diffusion and message passing
\citep{bodnar2022neuralsheaf,barbero2022sheaf,hajij2025copresheaftopologicalneuralnetworks}. Our contribution is to use exact Kan extensions in the
representable cases, and Kan-style structured extension more generally, as
a common language for attention, incidence mixing, learned topological
neighborhoods, and predict-detach completion.

\section{Discussion and Future Directions}
\label{sec:broader_impact}

The experiments support a more specific interpretation of KET than a
generic ``higher-order attention helps'' claim. In the strict-causal
setting, neighborhood design matters: quadratic KET is the strongest of
the compared causal models on WT2 and WT103, while the incidence
variant remains competitive with lower structural cost. Across all runs,
however, the dominant effect comes from the information regime rather
than from the neighborhood family alone. 

\subsection{Limitations and Future Directions}

Although we compare $12$ Transformer models and include depths \(L=8,16\)
and widths \(d=64,256\) in addition to the \(L=2,d=256\) headline
setting, the study remains modest in scale, is not a full factorial sweep,
and is centered on perplexity. We therefore view the main contribution
as a practical design lens and causal framing, not a final verdict on
large-scale architectures. Future work should scale neighborhood design
and information regime separately, testing richer simplicial families,
adaptive simplex selection, and longer-context sparse approximations.

Hybrid systems are another direction: KET provides prescribed
combinatorial structure, while TopoCoend learns geometric neighborhoods.
The structured-completion perspective also extends beyond the small
block-denoising setting: predict-detach, blockwise denoising, infilling,
and other partial-information objectives can be treated as neighboring
points in one design space.

\section{Conclusion}

We introduced KET as a structured-extension language for Transformer
sequence models in which a layer transports local information from chosen
source neighborhoods to target token positions. Attention is the
singleton-neighborhood case, incidence-style geometric mixing is a sparse
local case, and KET is the simplicial case. These updates are genuine
enriched Kan extensions under the representability and functoriality
conditions stated in the supplement; generic learned weighted sums are
Kan-style rather than automatically Kan extensions. Predict-detach blocks
an auxiliary gradient path and avoids direct transport of teacher-forced
hidden states, but strict autoregression additionally requires every carrier
to satisfy the target-relative filtration condition.

\section*{Supplementary Materials} 
\label{sec:appendix} 

\newpage
\section{Changes in This Version}
\label{app:changes}

This version incorporates three substantive corrections and clarifications.
\begin{enumerate}
\item \textbf{Causal status of predict-detach.}
Detaching a predictive carrier blocks gradients and avoids direct transport of
teacher-forced hidden states, but it does not by itself make a noncausal update
strictly autoregressive.  The revised paper gives a target-relative filtration
criterion: every carrier consumed at target \(t\) must be measurable with
respect to the prefix information available at \(t\).  Runs that mix
future-indexed predictive carriers are now described as noncausal
self-conditioning rather than strict autoregression.
\item \textbf{Scope of the Kan-extension interpretation.}
A generic learned weighted sum is no longer asserted to satisfy a Kan
universal property.  The supplement states explicit functoriality,
representability/naturality, and coend-realization conditions under which an
enriched update is an actual Kan extension.  Other learned aggregations are
called \emph{Kan-style}.
\item \textbf{Experimental scale and interpretation.}
The paper now states explicitly that the reported studies cover widths
\(d=64,256\) and depths \(L=2,8,16\).  A coverage table distinguishes the
headline language-model comparisons from the higher-depth structured-completion
runs, avoiding the implication of a full depth-by-width factorial sweep.
Cross-regime gains are attributed to the information regime rather than to
neighborhood geometry alone.
\end{enumerate}

We give a more detailed theoretical explanation for KET models here. We also provide
an accompanying supplementary reproducibility code repository that contains the
exact experiment harnesses, configs, logs, summaries, and checkpoints
used for the reported language-model comparisons and structured
completion pilots. 

\section{Kan Extensions as Universal Structured Extension}
\label{app:kan_extensions}

Machine learning research over the past six decades, ever since Gold's formalization of language identification in the limit \citep{GOLD1967447}, has been centered on the scientific principle of approximating functions over sets. 
Category theory \citep{maclane:71} fundamentally alters our perspective of machine learning, moving from functions to {\em functors}, structured mappings of categories that map both the {\em objects} and the {\em arrows} of an input category to an output category.  
Instead of investigating a smorgasboard of methods for an arbitrary extension of a function on points,
we instead redefine the problem in terms of an extension of a functor along a structural map.
Once the source values and the target structure are organized
functorially, universal constructions become available: 
\emph{left} and \emph{right Kan extensions} define principled ways to extend local information across a
larger domain relative to a specified structure.

This perspective is central to the architectures developed in this
paper.
A Transformer layer, a geometric mixing layer, or a higher-order
message-passing rule may all be viewed as structured extension
operators.
The key question is not merely \emph{which values are mixed}, but
\emph{along what structure} and according to \emph{which universal
principle} they are extended.

\subsection{Kan Extension Definition}

We summarize the main ideas underlying Kan Extensions, leaving the details to standard textbook treatments \citep{maclane:71,riehl2017category,richter2020categories}. 
Let
\[
F : \mathcal{C} \to \mathcal{E},
\qquad
K : \mathcal{C} \to \mathcal{D}
\]
be functors.
The functor \(F\) assigns values to objects of the source category
\(\mathcal{C}\), while \(K\) explains how \(\mathcal{C}\) sits inside,
or maps into, the larger target category \(\mathcal{D}\).
We would like to construct a functor
\[
H : \mathcal{D} \to \mathcal{E}
\]
that extends \(F\) along \(K\).

\begin{definition}
The \textbf{left Kan extension} of \(F\) along \(K\), denoted
\(\operatorname{Lan}_K F\), is a functor
\[
\operatorname{Lan}_K F : \mathcal{D} \to \mathcal{E}
\]
together with a natural transformation
\[
\eta : F \Rightarrow (\operatorname{Lan}_K F)\circ K
\]
such that for any other pair
\[
G : \mathcal{D} \to \mathcal{E},
\qquad
\gamma : F \Rightarrow G\circ K,
\]
there exists a unique natural transformation
\[
\alpha : \operatorname{Lan}_K F \Rightarrow G
\]
with
\[
\gamma = (\alpha K)\circ \eta.
\]
Dually, the \textbf{right Kan extension} \(\operatorname{Ran}_K F\)
is the universal extension characterized by maps
\[
\epsilon : (\operatorname{Ran}_K F)\circ K \Rightarrow F.
\]
\end{definition}

\begin{center}
\begin{tikzcd}[row sep=2cm, column sep=2.2cm]
\mathcal{C}
  \ar[r,"F"]
  \ar[d,"K"'] &
\mathcal{E} \\
\mathcal{D}
  \ar[ur,bend right=15,"G"']
  \ar[ur,bend left=15,"\operatorname{Lan}_K F", ""{name=LAN, below}] &
\arrow[Rightarrow, from=LAN, to=1-2, shorten <=7pt, shorten >=7pt, "\eta" description]
\end{tikzcd}
\end{center}

Succinctly, we can informally rephrase the definitions as:

\begin{quote}
A left Kan extension is the most universal way to \emph{build} values on
\(\mathcal{D}\) from data on \(\mathcal{C}\).  
A right Kan extension is the most universal way to \emph{complete}
values on \(\mathcal{D}\) so that they remain compatible with
\(\mathcal{C}\).
\end{quote}

For the purposes of this paper, the left Kan extension will play the
leading role, since it captures aggregation and contextualization.
The right Kan point of view will become increasingly important when we
turn to denoising, completion, and refinement.

\subsection{Pointwise Formulas: Colimits and Limits}

Kan extensions become especially intuitive when written pointwise.
For an object \(d \in \mathcal{D}\), the value
\((\operatorname{Lan}_K F)(d)\) is built from all source objects
\(c \in \mathcal{C}\) that map into \(d\).
Formally, one considers the comma category \((K \downarrow d)\),
whose objects are arrows
\[
Kc \to d
\]
in \(\mathcal{D}\), and whose morphisms are commuting triangles induced
by morphisms in \(\mathcal{C}\).

If the relevant colimits exist, then the left Kan extension is computed
pointwise by
\begin{equation}
(\operatorname{Lan}_K F)(d)
\;\cong\;
\operatorname*{colim}_{(K \downarrow d)} F.
\label{eq:pointwise_lan}
\end{equation}

Dually, the right Kan extension is built from all ways in which \(d\)
must be compatible with source objects.
Writing \((d \downarrow K)\) for the comma category whose objects are
arrows
\[
d \to Kc,
\]
one obtains the pointwise formula
\begin{equation}
(\operatorname{Ran}_K F)(d)
\;\cong\;
\operatorname*{lim}_{(d \downarrow K)} F.
\label{eq:pointwise_ran}
\end{equation}

These two formulas already capture the conceptual distinction that will
matter throughout the paper:

\begin{itemize}
    \item \(\operatorname{Lan}\) uses a \textbf{colimit}:
    it glues together many local contributions.
    \item \(\operatorname{Ran}\) uses a \textbf{limit}:
    it finds a jointly compatible value satisfying many constraints.
\end{itemize}

For machine learning readers, it is often enough to remember the
mnemonic:
\begin{quote}
left Kan = aggregation from evidence, \qquad
right Kan = completion under constraints.
\end{quote}

\subsection{Interpolation and Completion as Motivating Examples}

The difference between sets and categories can be seen in two familiar
problems.

\paragraph{Interpolation.}
Suppose we know values \(f(x_i)=y_i\) at a finite set of sample points.
As a set-theoretic problem, there is no canonical extension to all
query points \(x\): one may choose a constant predictor, a piecewise
linear interpolant, a polynomial, or a kernel smoother.
The ambiguity does not disappear in category theory, but it is
reframed.
One specifies a structural relation \(W(i,x)\) saying how each sample
\(x_i\) contributes to a query point \(x\), and the extension is then
computed relative to that structure.
In finite enriched settings, the resulting left Kan extension takes the
weighted form
\[
(\operatorname{Lan}_W F)(x)
\;\approx\;
\sum_i W(i,x)\,F(i).
\]
Thus the Kan extension is not a magical elimination of modeling
choices; it is the universal extension \emph{relative to the chosen
structure}.

\paragraph{Completion and denoising.}
Now suppose the problem is not to interpolate from samples but to
complete a partially observed or noisy state.
Again, many completions may be possible in principle.
What selects among them is the compatibility structure:
which local constraints must the completed state satisfy?
The right Kan viewpoint supplies an exact formalization only after the local
constraint diagram and its limiting universal property have been specified.
Instead of aggregating evidence from contributing sources, one computes
a value that is as compatible as possible with all the relevant local
views.
This is the categorical intuition behind denoising, structured
completion, and diffusion-style refinement.

These two examples foreshadow the main architectural claim of the
paper:
attention has the algebraic shape of a learned weighted left-Kan-style
aggregation, while
self-conditioned denoising and refinement point toward a right-Kan
interpretation.

\subsection{Coends as Weighted Aggregation}

The pointwise left Kan formula in
Eq.~\eqref{eq:pointwise_lan} can often be written in a more
computationally useful form using a \emph{coend}.
For readers unfamiliar with ends and coends, the essential intuition is
simple:
a coend behaves like a structured weighted sum, with the indexing
category enforcing the appropriate identifications.

The notation
\[
\int^{c} H(c,c)
\]
should be read informally as
\begin{quote}
``sum over all \(c\), while identifying contributions that are
equivalent under the morphisms of the indexing category.''
\end{quote}

In the settings relevant here, this behaves like a weighted pooling
operation.
If \(F(c)\) is a feature vector and \(W(c,t)\) is a compatibility
weight, then the coend has the computational flavor
\[
(\text{coend})(t)
\;\sim\;
\sum_c W(c,t)\,F(c),
\]
possibly with identifications imposed by the structure of the category.

This is the computational pattern that appears in attention, geometric
message passing, and Kan-style pooling:
values are aggregated from many structured sources, and the weights
encode how strongly each source contributes to the target.

\subsection{Kan Extensions via Coends}

A standard enriched formula expresses the pointwise left Kan extension
as a coend:
\begin{equation}
(\operatorname{Lan}_K F)(d)
\;\cong\;
\int^{c \in \mathcal{C}}
\mathcal{D}(Kc,d)\,\otimes\,F(c).
\label{eq:kan_coend}
\end{equation}

Here \(\mathcal{D}(Kc,d)\) is the hom-object measuring how \(Kc\)
contributes to \(d\), and \(\otimes\) denotes the enrichment, which in
our neural setting behaves like scalar weighting of feature vectors.
Equation~\eqref{eq:kan_coend} therefore says:

\begin{quote}
To compute the value at \(d\), gather contributions from every source
object \(c\), weight them by how strongly \(Kc\) points to \(d\), and
aggregate them universally.
\end{quote}

\subsection{When the Neural Update Is Actually a Kan Extension}
\label{sec:kan_exactness}

The coend formula gives a precise criterion, not a license to call every
weighted sum a Kan extension. Let \(\mathcal V\) be a symmetric monoidal
closed enrichment, let \(\mathcal E\) admit the required weighted colimits,
and let
\[
K:\mathcal C\to\mathcal D,
\qquad
F:\mathcal C\to\mathcal E
\]
be enriched functors. A neural update at \(d\) is the enriched left Kan
extension \((\operatorname{Lan}_KF)(d)\) when all of the following hold:
\begin{enumerate}
\item source features form the functor \(F\);
\item the weights are the representable hom-objects
      \(W(c,d)\cong\mathcal D(Kc,d)\), naturally in \(c\) and \(d\);
\item the implemented aggregation realizes the tensor and coend quotient in
      Eq.~\eqref{eq:kan_coend}.
\end{enumerate}
Under these assumptions the universal property follows from the standard
enriched coend formula.

For an arbitrary learned kernel \(W(c,d)\), the expression
\[
\int^c W(c,d)\otimes F(c)
\]
is a profunctor-weighted transform. It is a Kan extension along a functor
\(K\) only in the special case that \(W\) is representable as
\(\mathcal D(K-,d)\) (or when an explicitly enlarged category makes the
learned kernel representable and natural). Normalization by softmax does not
establish representability, functoriality, or the universal comparison
property. Consequently, the generic Transformer and learned-neighborhood
updates in this paper are called \emph{Kan-style}; the exact Kan claim is
reserved for implementations satisfying the three conditions above.

Dually, in enriched settings the right Kan extension may be written in
end form as
\begin{equation}
(\operatorname{Ran}_K F)(d)
\;\cong\;
\int_{c \in \mathcal{C}}
[\mathcal{D}(d,Kc),\,F(c)],
\label{eq:ran_end}
\end{equation}
where the end expresses compatibility with all incoming constraints.
We will not need this formula computationally in most of the paper,
but it is conceptually important: left Kan extensions are naturally
coend-like and aggregation-oriented, while right Kan extensions are
end-like and compatibility-oriented.

\subsection{Interpretation for Transformer Architectures}

Equation~\eqref{eq:kan_coend} has a direct neural interpretation.

\begin{itemize}
    \item The source objects \(c\) correspond to tokens, edges, or
    higher-order simplices.
    \item The target object \(d\) is typically a token position whose
    representation is being updated.
    \item The hom-objects \(\mathcal{D}(Kc,d)\) play the role of
    compatibility or neighborhood weights.
    \item The values \(F(c)\) are the feature vectors carried by the
    source objects.
\end{itemize}

Under this interpretation, a familiar update
\[
h'_d
=
\sum_c w_{d,c}\,V_c
\]
has the algebraic shape of a discretized enriched coend computation. It is
precisely a Kan extension only under the criterion in
Section~\ref{sec:kan_exactness}.
The architectural families in this paper can then be read in a common
language:

\begin{itemize}
    \item \textbf{Transformers} aggregate over token neighborhoods;
    attention is a weighted left-Kan-style pooling rule on a sequence.
    \item \textbf{Kan Extension Transformers (KET)} make the source
    category explicit, lifting from tokens to simplices and then
    extending back to token updates via simplicial incidence.
    \item \textbf{TopoCoend Transformers} retain the weighted coend
    form but define the neighborhood category through learned
    topological geometry rather than fixed simplicial structure.
\end{itemize}

Thus KET and TopoCoend should not be viewed as unrelated alternatives.
They are two realizations of the same broader design idea:
structured contextualization as weighted extension. Universality is an
additional property of the representable cases.

\subsection{Why This Matters for the paper}

This section provides the conceptual bridge from ordinary geometric
mixing to the models introduced later.

\begin{itemize}
    \item It explains why attention and geometric aggregation can be
    treated uniformly as structured extension operators.
    \item It clarifies why KET naturally motivates the language of Kan
    extensions: the model transports values from simplicial structure back
    to tokens, and exact instances satisfy
    Section~\ref{sec:kan_exactness}.
    \item It explains why TopoCoend lives naturally in the language of
    coends: the model aggregates over learned neighborhoods using
    weighted geometric compatibility.
    \item It prepares the later analysis of information regimes by
    distinguishing aggregation from compatibility and extension from
    leakage.
\end{itemize}

The main lesson is that Kan extensions and coends provide a useful vocabulary
for describing how local token information is transported, aggregated, and
reorganized across structured domains. The universal vocabulary applies
literally in the exact cases above; elsewhere it is a disciplined analogy
that records source objects, targets, and weights without asserting a
universal property.

\section{From Autoregression to Structured Completion}

In the categorical view of Transformer-style
architectures presented in this paper, attention, geometric mixing, simplicial
aggregation, and topological neighborhood pooling all appear as
instances of structured extension operators.
Kan Extension Transformers (KET) make this explicit: a layer can be
understood not merely as a hand-designed mixing rule, but as a
principled way of extending local token representations across a richer
neighborhood system.
Our experiments also highlight a central empirical fact:
\emph{information regime often matters more than architectural detail}.
In light of this, we can ask a more fundamental question: 

\begin{quote}
What is the right \emph{prediction regime} for language modeling once we
stop assuming that generation must proceed one token at a time?
\end{quote}

In standard autoregressive language modeling, the model learns a map
\begin{equation}
f : C_t \longrightarrow \Sigma,
\end{equation}
where \(C_t\) is the causal context at position \(t\) and \(\Sigma\) is
the vocabulary object.
This formulation has been extraordinarily successful, but it builds in a
strong modeling choice:
the future is generated as a sequence of isolated single-token decisions.
From the categorical perspective,
there is nothing inevitable about this restriction.
Autoregression is one particular way of extending a local context into a
future prediction, but it is not the only one.

More generally, language modeling may be viewed as a problem of
\emph{structured extension} or \emph{structured completion}.
Instead of predicting one token, a model may attempt to predict an
entire future block at once, or it may attempt to complete a partially
specified future structure.
These two possibilities correspond to different information regimes:

\begin{itemize}
\item \textbf{Direct block prediction:}
learn a map
\[
C_t \longrightarrow \Sigma^B,
\]
which predicts a block of \(B\) future tokens in one shot.

\item \textbf{Denoising block completion:}
learn a map
\[
(C_t,\tilde{x}_{t+1:t+B}) \longrightarrow x_{t+1:t+B},
\]
which completes a corrupted or partially specified future block.
\end{itemize}

The difference between these regimes is not merely cosmetic.
Direct block prediction asks the model to construct an entire structured
future from context alone.
Denoising instead treats the future block as a partially specified
object and asks the model to complete it compatibly.
The first regime is closer to a
one-shot extension problem, while the second is closer to a structured
completion problem. We prefer the latter formulation in this paper. 

From the KET perspective, this shift is especially meaningful.
We  emphasized above that left Kan extensions as a principled
language for aggregation and contextualization.
Here we begin to see the complementary side of the story:
language modeling is not only about aggregating evidence from the past,
but also about completing a future structure under internal
compatibility constraints.
Autoregressive next-token prediction, direct block prediction, and
denoising completion should therefore be seen as distinct
\emph{information regimes} for the same underlying structured prediction
problem.

The central empirical claim of this paper is that these information
regimes differ dramatically in difficulty and effectiveness.
Across datasets and architectures, direct block prediction proves much
harder than denoising completion, and denoising often yields much lower
perplexity than direct block generation.
This suggests that language modeling is better understood not merely as
one-step prediction, but as a problem of \emph{structured completion}.

Kan-style architectures tell us how to extend and organize information
over structured domains.
This paper asks what kind of \emph{target structure} should be
predicted.
Once that question is posed, the design space widens substantially:
the target need not be a single token, but may be a block, a corrupted
future configuration, or a partially specified linguistic object.
The choice among these regimes turns out to be one of the most important
determinants of empirical performance.

\section{Direct Block Prediction as a Product-Object Target}

A first generalization of standard autoregressive prediction is to
predict not a single next token, but an entire future block at once.
Instead of producing only \(x_{t+1}\), the model predicts
\[
x_{t+1:t+B},
\]
where \(B\) is the block size.

The corresponding prediction map is
\begin{equation}
f : C_t \longrightarrow \Sigma^B,
\end{equation}
where \(C_t\) is the causal context at position \(t\) and
\(\Sigma^B\) denotes the \(B\)-fold product of the vocabulary object:
\[
\Sigma^B
=
\Sigma \times \Sigma \times \cdots \times \Sigma .
\]

Categorically, direct block prediction therefore replaces the usual
single-token target \(\Sigma\) by a structured \emph{product object}.
The model must map the context \(C_t\) into a tuple of future tokens in
one step rather than constructing that future sequentially.

This formulation is natural, but it is substantially harder than
ordinary next-token prediction.
The target space now has cardinality \(|\Sigma|^B\), so the model must
select a coherent block from a combinatorially larger output space.
Equivalently, it must infer not only which tokens should appear, but
which \emph{joint configuration} of tokens is compatible with the
context.

Direct block prediction is
still an extension problem, but it is a particularly demanding one:
the model must extend the causal context directly into a fully specified
future structure, without any partial information about that structure
being provided as input.
This is precisely what distinguishes it from denoising completion, where
the future block is only partially specified and the task becomes one of
structured refinement rather than one-shot generation.

In practice, our experiments show that direct block prediction performs
surprisingly poorly even when paired with strong architectures.
This suggests that the main difficulty is not architectural weakness,
but the intrinsic hardness of predicting a coherent product-structured
target in a single step.

\section{Denoising as Structured Completion}

A different way to change the language-modeling objective is to change
not the architecture, but the \emph{information regime}.
Instead of predicting the future block from context alone, we provide
the model with a corrupted or partially specified version of that block
and ask it to reconstruct the original:
\begin{equation}
f : (C_t,\tilde{x}_{t+1:t+B}) \longrightarrow x_{t+1:t+B},
\label{eq:denoise_map}
\end{equation}
where \(\tilde{x}_{t+1:t+B}\) is a noisy, masked, or otherwise corrupted
version of the future block.

This changes the task in an important way.
Direct block prediction asks the model to generate a fully specified
future object from scratch.
Denoising instead presents the future as a \emph{partially specified
structure} and asks the model to complete it.
Some coordinates are already present, others are uncertain or corrupted,
and the model must infer a coherent full block compatible with both the
causal context and the partial future evidence.

From a categorical perspective, this is closer to a
\emph{completion problem} than to a one-shot generation problem.
The corrupted block already carries partial structure, and the model
learns to extend that partial object to a full consistent one.
This suggests a
complementary viewpoint to left-Kan aggregation:
where left Kan extensions capture the aggregation of evidence from local
sources, denoising naturally points toward a \emph{right-Kan intuition},
in which the task is to find a value or structure that is compatible
with multiple local constraints at once.

This perspective helps explain why denoising is often easier than direct
block prediction.
The model is not forced to choose a block in \(\Sigma^B\) from context
alone.
Instead, it solves a constrained completion problem in which part of the
future structure is already given.
Empirically, this distinction turns out to matter greatly:
denoising block prediction consistently outperforms direct block
prediction across the architectures studied in this paper.

For this reason, the experiments below suggest that language modeling is
often better viewed as a problem of \emph{structured completion} rather
than pure one-shot generation.
\section{Simplicial Structure and Horn Completion}

The denoising regime admits a natural simplicial interpretation.
A corrupted future block is not merely noisy data; it is a
\emph{partially specified structured object}.
Some components of the future are given, others are masked, corrupted,
or uncertain, and the model is asked to infer a coherent completion.
This makes denoising fundamentally different from direct block
prediction.
The task is no longer to generate a full product-structured target from
scratch, but to complete an already partially present configuration.

A useful geometric analogy is provided by simplicial homotopy theory.
A horn is a partial simplex in which enough faces are present to specify
most of the structure, but one face or relation remains missing.
To fill the horn is to extend this partial object to a full simplex.
A simplicial set is called a Kan complex if every horn can be filled
\cite{may1992simplicial}.
We do not claim that a denoising language model literally defines a Kan
complex in the strict homotopy-theoretic sense.
Rather, the horn-filling picture provides the right structural
intuition:
the model receives a partial future object and must construct a
completion compatible with the local information already present.

This perspective is especially natural from the KET viewpoint developed
in this paper.
Left Kan extensions organized the aggregation of contextual
evidence from structured neighborhoods.
Here, denoising introduces the complementary intuition of
\emph{compatibility-based completion}.
The corrupted block already imposes constraints on the space of valid
futures, and the model must fill in the missing coordinates in a way
that respects both the causal context \(C_t\) and the internal
relations within the block itself.
In this sense, denoising is closer to a right-Kan-style completion
problem than to a one-shot left-Kan-style aggregation problem.

The horn-filling view also helps explain the empirical advantage of
denoising over direct block prediction.
Direct block prediction asks the model to select a point in the full
product object \(\Sigma^B\) from context alone.
Denoising, by contrast, constrains the search space by presenting a
partial object whose completion must satisfy local compatibility
conditions.
This additional structure makes the task easier and more stable.
Empirically, as the experiments  show, that difference in
information regime matters far more than small architectural variations.

For this reason, denoising language modeling is best viewed not simply
as another decoding trick, but as a structured completion problem with a
natural simplicial interpretation.
The horn-filling picture makes precise the sense in which the model is
asked to complete a partial future rather than emit a future from
nothing.

\section{Partial Information, Horn Filling, and Structured Completion}

The contrast between direct block prediction and denoising block
prediction may now be stated more precisely.

In direct block prediction, the model learns a morphism
\[
f : C_t \longrightarrow \Sigma^B,
\]
mapping a causal context \(C_t\) directly into the \(B\)-fold product
object of future tokens.
The target is a fully specified block, and the model must select a
coherent point of \(\Sigma^B\) in a single step.
Categorically, the prediction problem is therefore posed entirely in
terms of a map from context to a complete structured target.

In denoising block prediction, by contrast, the model is given a
corrupted or partially specified future block \(\tilde{x}\) and learns a
map
\[
f : (C_t,\tilde{x}) \longrightarrow x_{t+1:t+B}.
\]
The crucial difference is that the domain now contains \emph{partial
information about the target object itself}.
The model is not asked to choose a future block from scratch.
Instead, it must complete a partially specified structure in a way that
is compatible with both the context and the information already present
in \(\tilde{x}\).

This distinction can be expressed diagrammatically as
\[
\begin{tikzcd}
C_t \arrow[r] & \Sigma^B
\end{tikzcd}
\qquad\text{versus}\qquad
\begin{tikzcd}
(C_t,\tilde{x}) \arrow[r] & \Sigma^B .
\end{tikzcd}
\]
The codomain is the same in both cases, but the domain is richer in the
denoising regime.
That extra structure dramatically changes the learning problem.
Direct block prediction searches over the full product object, whereas
denoising searches over a constrained family of completions consistent
with a partial specification.

\subsection{Horn Filling as a Structural Analogy}

This completion process admits a natural simplicial analogy.
A block of tokens may be viewed as a simplex whose vertices correspond
to token positions and whose internal relations encode coherence across
the block.
A corrupted block \(\tilde{x}\) then behaves like a partially specified
simplex: some vertices or relations are present, while others are
missing or uncertain.

In simplicial homotopy theory, such partial simplices are often
described as \emph{horns}.
To fill a horn is to extend a partial simplex to a full simplex.
A simplicial set is called a Kan complex when every horn admits such a
filler.
We do not claim that a denoising language model literally constructs a
Kan complex in the strict homotopy-theoretic sense.
Rather, the horn-filling picture provides the correct structural
intuition:
the model is given a partial future object and must construct a coherent
completion.

This perspective sharpens the empirical lesson of this paper.
Denoising is easier than direct block prediction not merely because it is
``given more input,'' but because it solves a more structured problem.
The corrupted block already narrows the space of admissible futures, and
the model must fill in what is missing rather than generate everything
from nothing.

\subsection{Relation to KET}

KET and related architectures
provide a principled language for \emph{structured extension}:
local token information is extended across simplicial or topological
neighborhood systems by universal aggregation mechanisms.
We are in a sense generalizing attention from the structure of the
\emph{aggregator} to the structure of the \emph{target}.

From this perspective:
\[
C_t \to \Sigma
\]
is ordinary next-token prediction,
\[
C_t \to \Sigma^B
\]
is direct prediction of a product-structured future,
and
\[
(C_t,\tilde{x}) \to x_{t+1:t+B}
\]
is structured completion in a category of partial information.

This is the deeper connection to the broader program of the book.
Intelligence is often not best understood as one-shot emission of fully
specified outputs, but as the ability to extend partial structures into
coherent global structures within an appropriate category of
information.
Under that reading, denoising language modeling is not a minor variant
of decoding.
It is a concrete instance of a much more general principle of
structured completion.

\section{Predict--Detach as a Modal Boundary}
\label{sec:predict_detach_category}

The predict--detach mechanism plays a central role in the valid
self-conditioned regimes studied in this paper.
Operationally, it is simple: the model computes a predictive carrier and
then applies \texttt{detach()} so that the carrier may participate in
forward computation without transmitting gradients backward through the
predictive branch.
Conceptually, however, this is more than an implementation trick.
It introduces a separation between two kinds of structure:
the \emph{forward semantics} of the model and the \emph{learning
semantics} induced by reverse-mode differentiation.

This section gives a categorical formulation of that separation.
The key idea is that predict--detach acts as a \emph{modal boundary}:
it preserves forward information flow while annihilating the associated
gradient flow.
This viewpoint makes precise what detach does to optimization. It does not,
by itself, prove that a noncausal forward aggregation is free of
future-token information; that requires the separate target-relative
filtration test.

\subsection{Forward and Backward Semantics}

A neural layer has two mathematically distinct aspects.

\paragraph{Forward semantics.}
A forward operator transforms representational states:
\[
f : X \to Y,
\]
where \(X\) and \(Y\) may be token sequences, hidden states,
predictive distributions, or embedding carriers.

\paragraph{Backward semantics.}
Under reverse-mode automatic differentiation, the same operator induces
a backward map on cotangent or gradient objects:
\[
f^\ast : \partial Y \to \partial X.
\]

Thus every learned computation carries both a forward action and a
backward action.
For the purposes of this paper, it is convenient to package these
together into a paired semantics
\[
f \mapsto (f, f^\ast).
\]

\subsection{Detach as a Modal Operator}

The detach operation preserves the forward component of an operator but
kills its backward component.

\begin{definition}[Detach modality]
For an operator with paired semantics \((f,f^\ast)\), define
\[
\Box(f,f^\ast) := (f,0).
\]
Equivalently, \(\Box\) leaves the forward map unchanged while replacing
the backward map by the zero morphism.
\end{definition}

This formalizes the implementation-level meaning of
\texttt{detach()}:
the carrier remains available to subsequent layers as a forward signal,
but it cannot serve as a path for gradient propagation.

\paragraph{Interpretation.}
The modality \(\Box\) is not a change to what the model computes in the
forward pass.
It is a change to which paths are visible to learning.
In this sense, predict--detach imposes a modal boundary between
\emph{semantic use} and \emph{gradient credit assignment}.

\subsection{Predict--Detach as a Composite}

Let
\[
\mathrm{Pred} : H \to E
\]
denote the predictive carrier map from hidden states \(H\) to embedding
space \(E\):
\[
\mathrm{Pred}(h)
=
\operatorname{softmax}(Wh/T)\,E.
\]

The predict--detach carrier is then the composite
\[
\mathrm{PredDetach}
=
\Box \circ \mathrm{Pred}.
\]

Thus a hidden state first generates a predictive embedding and that
embedding is then passed into later geometric or simplicial aggregation
only through the detached channel:
\[
H \xrightarrow{\mathrm{Pred}} E \xrightarrow{\Box} E.
\]

In paired semantics, this becomes
\[
(\mathrm{Pred},\mathrm{Pred}^\ast)
\quad\mapsto\quad
(\mathrm{Pred},0).
\]

So the predictive carrier participates in forward aggregation, but the
predictive branch itself does not receive gradient contributions from
that later aggregation.

\subsection{Information Regimes Revisited}

This modal viewpoint clarifies the three information regimes used
throughout the paper.

\paragraph{Strict causal regime.}
All forward morphisms respect temporal order.
No future-position object contributes to the update at time \(t\).
This is the ordinary autoregressive regime.

\paragraph{Gold noncausal regime (invalid).}
Future-position hidden states computed under teacher forcing are allowed
to contribute directly to the update at time \(t\).
Since those hidden states encode gold future tokens, this introduces an
illegitimate information path.

\paragraph{Predict--detach regime.}
Future-position carriers may appear structurally in the neighborhood
system, but only through predictive values passed through the detach
modality.
Thus the forward computation may exploit richer noncausal structure,
while the learning dynamics remain shielded from direct gradient flow
through that channel.

The essential distinction is therefore not just whether the neighborhood
graph is noncausal.
It is both whether the channel carries a hidden state or a prediction and
whether that prediction is measurable from the prefix available to the
target receiving it. Detach changes neither forward values nor their
measurability.

\subsection{Causal Validity Requires Target-Relative Measurability}

Predict--detach is causally legitimate for a target only when the carrier is
generated from that target's available prefix and is then insulated from
backward influence.

Let
\[
\hat e_t
=
\operatorname{detach}\!\Bigl(
\operatorname{softmax}(\ell_t/T)\,E
\Bigr),
\qquad
\ell_t = W_o h_t.
\]
Since \(h_t\) is causal, \(\hat e_t\) is a function only of
\(x_{\le t}\) and is admissible for predicting \(x_{t+1}\). More generally,
a carrier \(\hat e_s\) consumed by the update at \(t\) is admissible only if
it is measurable with respect to \(\mathcal F_t=\sigma(x_{\le t})\).
A carrier computed at \(s>t\) may depend on \(x_{t+1:s}\), which is
gold-future information relative to target \(t\), even though the carrier
is a prediction and is detached.

The shifted-previous construction
\(\detach(p_{t-1}E)\) satisfies this test. A same-index carrier passed
through a symmetric mixer can reach earlier targets and therefore does not
generally satisfy it. The gold-noncausal regime fails more directly because
it transports teacher-forced future hidden states themselves.

\subsection{Diagrammatic View}

The contrast between an ordinary predictive branch and a detached one can
be summarized diagrammatically as
\[
\begin{tikzcd}[column sep=large,row sep=large]
H \arrow[r,"\mathrm{Pred}"] \arrow[d]
& E \arrow[d] \\
\partial H
& \partial E \arrow[l,"\mathrm{Pred}^\ast"']
\end{tikzcd}
\qquad\leadsto\qquad
\begin{tikzcd}[column sep=large,row sep=large]
H \arrow[r,"\mathrm{PredDetach}"] \arrow[d]
& E \arrow[d] \\
\partial H
& \partial E \arrow[l,"0"']
\end{tikzcd}
\]

The left diagram represents an ordinary predictive carrier with both
forward and backward semantics.
The right diagram represents the detached carrier:
the forward morphism remains, while the backward morphism is zeroed out.

\subsection{Relation to KET and TopoCoend}

This modal interpretation applies uniformly across the models of the
paper.

\begin{itemize}
    \item In \textbf{quadratic KET}, detached predictive carriers may be
    lifted to simplex values and then aggregated globally.
    \item In \textbf{incidence-restricted KET}, detached carriers may be
    passed along local simplicial incidence relations.
    \item In \textbf{TopoCoend}, detached carriers may be embedded into a
    learned topological coordinate system and used to define fuzzy
    neighborhood aggregation.
\end{itemize}

Thus predict--detach is not specific to one architecture.
It is a general device for permitting richer structural forward computation
without transmitting gradients through the predictive branch. Whether that
forward computation is strictly autoregressive is a separate indexing
property.

\subsection{Conceptual Summary}

The main point of this section is that predict--detach should not be
viewed merely as a framework-specific trick.
It has a clean categorical meaning:

\begin{quote}
predict--detach is a modal boundary that preserves forward semantic
carriers while discarding their backward learning morphisms.
\end{quote}

This perspective explains how self-conditioned aggregation separates
forward reuse from backward credit assignment and why the same device can
be used uniformly across GT, KET, and TopoCoend architectures. Only
filtration-preserving instances are claimed to be strictly autoregressive.

\section{Information Regimes in Geometric Transformers}
\label{app:info_regimes}

In this section, we study two questions: \emph{architecture class} and
\emph{information regime}. A model may keep causal self-attention while
receiving an auxiliary side channel through a geometric branch. We classify
comparisons by whether every auxiliary morphism factors through the
target-relative prefix (strict causal regime C), uses detached endogenous
self-predictions without guaranteeing that factorization (self-conditioned
regime E), or explicitly introduces future-informative paths
(augmented-context regime A).

\bigskip

\subsection{Categorical View of Information Regimes}

Let $\mathcal{I}_{\le t}$ denote the prefix index category up to position
$t$. A strict autoregressive model is a functor over
$\mathcal{I}_{\le t}$, and all update paths to $\hat y_t$ factor through
prefix objects (Regime C). Auxiliary augmentation is represented by an added
morphism $\alpha_t:A_t\to H_t$. If $\alpha_t$ is provided by an external
future-informative channel, it defines Regime A. If it uses an endogenous
prediction and detach but its target-relative factorization is not enforced,
we label it Regime E rather than C. We analyze six practical variants:
\begin{itemize}
\item \texttt{transformer\_causal}
\item \texttt{transformer\_future\_hint}
\item \texttt{gt\_causal}
\item \texttt{gt\_noncausal}
\item \texttt{gt\_pred\_next\_detach}
\item \texttt{gt\_pred\_prev\_causal\_detach}
\end{itemize}
and report each one with an explicit regime label and side-channel description.

\paragraph{FutureHint vs PredictiveHint.}
\texttt{transformer\_future\_hint} and
\texttt{gt\_pred\_next\_detach} can produce similar trajectories, but they
are not the same mechanism. The former is an exogenous hint pathway (A);
the latter is endogenous self-conditioning (E). Its hint is generated by
the model and detached, but a symmetric mixer can transport a carrier
generated from a later source prefix into an earlier target. Regime E is
therefore reported separately and is not called strict autoregression.

\subsection{Kan Extension Transformer as Self-Conditioned Completion}
\label{sec:kando-kan-transformer}

Let $\mathcal I_{\le t}$ be the prefix index category and $F_{\le t}:\mathcal I_{\le t}\to\mathbf{Vect}$ the prefix-state functor.
At step $t$, we use the following schematic description of two coupled
operations:
\[
\widehat{x}_{t+1} \;\approx\; \mathrm{Ran}_{\iota_t}(F_{\le t})(t+1),
\]
and
\[
h_t^{\mathrm{geo}}
\;\approx\;
\operatorname*{colim}_{\mathcal I_{\le t}\cup\{t+1\}}
\Big(F_{\le t}\ \cup\ \widehat{x}_{t+1}\Big),
\]
where $\iota_t:\mathcal I_{\le t}\hookrightarrow \mathcal I_{\le t+1}$ is the canonical inclusion.
The first line is a right-Kan-style predictive completion (a next-token
estimate from prefix information), while the second is a
left-Kan/colimit-style aggregation that reuses this completion as an
auxiliary object. These displayed approximations are not asserted to be
actual Kan extensions unless their predictor and mixer satisfy
Section~\ref{sec:kan_exactness}.

Operationally, this yields a \emph{self-conditioned structured-extension
transformer}:
\begin{enumerate}
\item predict a next-token distribution from prefix-derived hidden states;
\item map this prediction into an embedding-level auxiliary branch;
\item aggregate hidden and auxiliary streams through the geometric mixer.
\end{enumerate}

This explains the observed transition regime of predictive-hint GT variants.
Early in training, $\widehat{x}_{t+1}$ is noisy and behavior remains close to strict-causal controls.
As prediction quality improves, the induced geometric update approaches the
behavior of augmented-context GT variants. In this sense, the model
interpolates between strict-causal and augmented-context behavior through an
endogenous predictive channel. This is a behavioral description, not a
claim that the noncausal variant is strictly autoregressive or an exact Kan
completion.

We distinguish three regimes:
\begin{enumerate}
\item strict causal: no auxiliary completion channel;
\item self-conditioned predictive completion: auxiliary channel built from $\widehat{x}_{t+1}$;
\item exogenous augmented channel: auxiliary channel uses externally provided future-informative signal.
\end{enumerate}

\subsection{The Geometric Transformer}
\label{sec:wikitext-gt}


\tikzset{
  block/.style={draw, rounded corners, minimum width=2.8cm,
                minimum height=0.85cm, align=center, font=\footnotesize},
  attn/.style={block, fill=orange!20},
  ffn/.style={block, fill=blue!15},
  norm/.style={block, fill=yellow!20},
  geom/.style={block, fill=green!18},
  hint/.style={block, fill=cyan!18},
  warn/.style={block, fill=red!12},
  merge/.style={circle, draw, inner sep=1.2pt, fill=white},
  io/.style={font=\footnotesize, align=center},
  encbox/.style={draw, rounded corners, inner sep=4pt},
  flow/.style={-Stealth, thick},
  skip/.style={-Stealth, thick},
}

The Geometric Transformer augments Transformer-style global mixing with a geometric branch defined over local simplicial structure. In sequence tasks, token indices are $0$-simplices and adjacent pairs form the $1$-skeleton. The key update in this paper is to separate \emph{architecture} from \emph{information budget} by explicitly modeling causal versus augmented-context variants. The architectural diagram on the next page clarifies this distinction: (a) Baseline Transformer (b)  GT with the earlier (non-causal) geometric mixer (c) GT-augmented  (d) Predicted-hint GT.

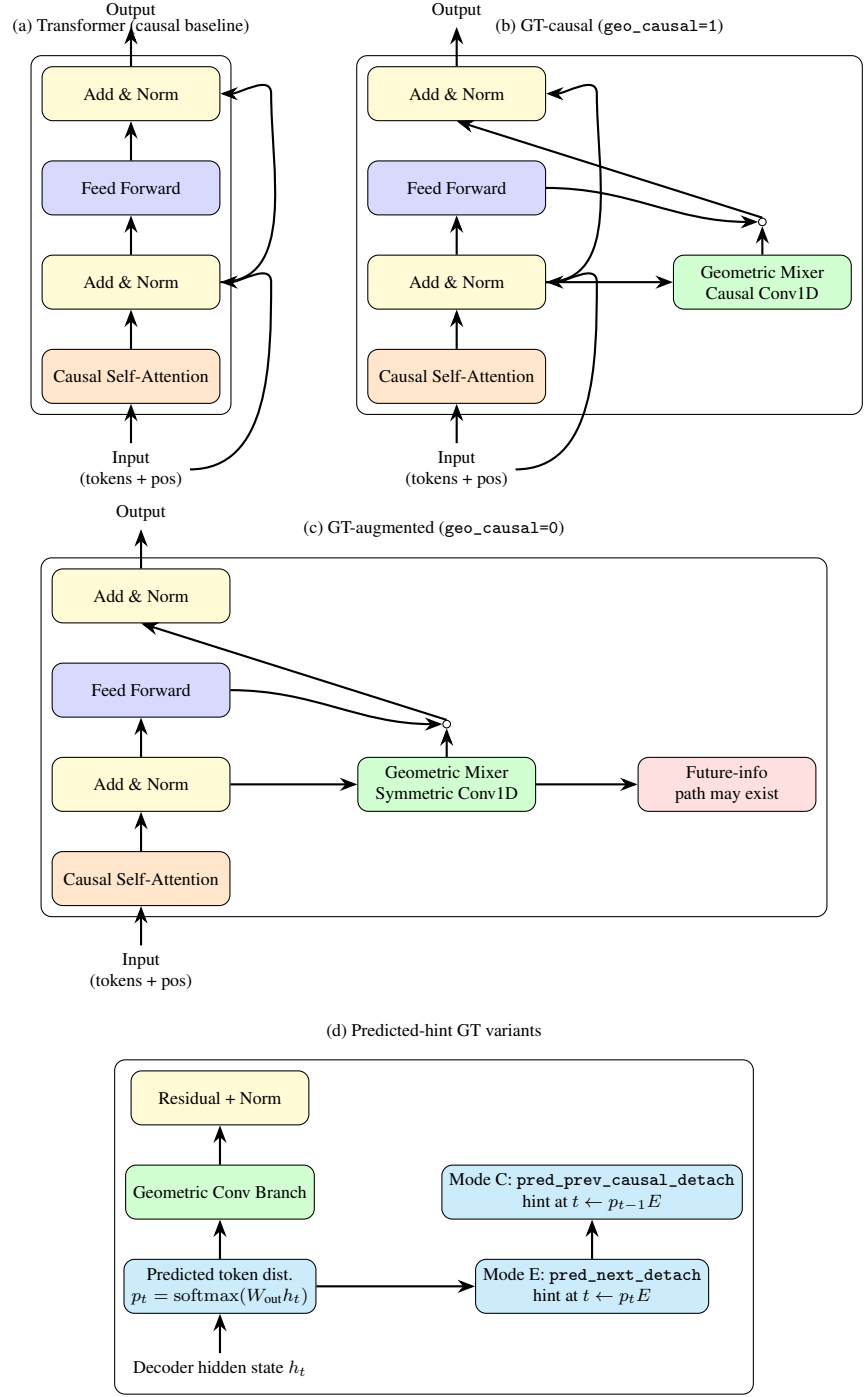
\begin{figure*}[p]
\centering
\begin{tikzpicture}[node distance=0.62cm,scale=0.84,transform shape]
  \node[io] (in_a) {Input\\(tokens + pos)};
  \node[attn, above=of in_a] (attn_a) {Causal Self-Attention};
  \node[norm, above=of attn_a] (n1_a) {Add \& Norm};
  \node[ffn, above=of n1_a] (ffn_a) {Feed Forward};
  \node[norm, above=of ffn_a] (n2_a) {Add \& Norm};
  \node[io, above=of n2_a] (out_a) {Output};

  \draw[flow] (in_a) -- (attn_a);
  \draw[flow] (attn_a) -- (n1_a);
  \draw[flow] (n1_a) -- (ffn_a);
  \draw[flow] (ffn_a) -- (n2_a);
  \draw[flow] (n2_a) -- (out_a);

  \draw[skip] (in_a.east) to[out=0,in=-90] ([xshift=0.8cm]n1_a.east) to[out=90,in=0] (n1_a.east);
  \draw[skip] (n1_a.east) to[out=0,in=-90] ([xshift=0.8cm]n2_a.east) to[out=90,in=0] (n2_a.east);

  \node[encbox, fit=(attn_a) (n2_a), label={[yshift=0.2cm]\footnotesize (a) Transformer (causal baseline)}] {};
\end{tikzpicture}
\hspace{0.6cm}
\begin{tikzpicture}[node distance=0.62cm,scale=0.84,transform shape]
  \node[io] (in_b) {Input\\(tokens + pos)};
  \node[attn, above=of in_b] (attn_b) {Causal Self-Attention};
  \node[norm, above=of attn_b] (n1_b) {Add \& Norm};
  \node[ffn, above=of n1_b] (ffn_b) {Feed Forward};
  \node[geom, right=2.0cm of n1_b] (geom_b) {Geometric Mixer\\Causal Conv1D};
  \node[merge, above=0.45cm of geom_b] (m_b) {};
  \node[norm, above=of ffn_b] (n2_b) {Add \& Norm};
  \node[io, above=of n2_b] (out_b) {Output};

  \draw[flow] (in_b) -- (attn_b);
  \draw[flow] (attn_b) -- (n1_b);
  \draw[flow] (n1_b) -- (ffn_b);
  \draw[flow] (ffn_b.east) to[out=0,in=180] (m_b.west);
  \draw[flow] (n1_b) -- (geom_b);
  \draw[flow] (geom_b) -- (m_b);
  \draw[flow] (m_b.north) -- (n2_b.south);
  \draw[flow] (n2_b) -- (out_b);

  \draw[skip] (in_b.east) to[out=0,in=-90] ([xshift=0.8cm]n1_b.east) to[out=90,in=0] (n1_b.east);
  \draw[skip] (n1_b.east) to[out=0,in=-90] ([xshift=0.8cm]n2_b.east) to[out=90,in=0] (n2_b.east);

  \node[encbox, fit=(attn_b) (n2_b) (geom_b), label={[yshift=0.2cm]\footnotesize (b) GT-causal (\texttt{geo\_causal=1})}] {};
\end{tikzpicture}
\hspace{0.6cm}
\begin{tikzpicture}[node distance=0.62cm,scale=0.84,transform shape]
  \node[io] (in_c) {Input\\(tokens + pos)};
  \node[attn, above=of in_c] (attn_c) {Causal Self-Attention};
  \node[norm, above=of attn_c] (n1_c) {Add \& Norm};
  \node[ffn, above=of n1_c] (ffn_c) {Feed Forward};
  \node[geom, right=2.0cm of n1_c] (geom_c) {Geometric Mixer\\Symmetric Conv1D};
  \node[warn, right=1.6cm of geom_c] (leak_c) {Future-info\\path may exist};
  \node[merge, above=0.45cm of geom_c] (m_c) {};
  \node[norm, above=of ffn_c] (n2_c) {Add \& Norm};
  \node[io, above=of n2_c] (out_c) {Output};

  \draw[flow] (in_c) -- (attn_c);
  \draw[flow] (attn_c) -- (n1_c);
  \draw[flow] (n1_c) -- (ffn_c);
  \draw[flow] (ffn_c.east) to[out=0,in=180] (m_c.west);
  \draw[flow] (n1_c) -- (geom_c);
  \draw[flow] (geom_c) -- (m_c);
  \draw[flow] (m_c.north) -- (n2_c.south);
  \draw[flow] (n2_c) -- (out_c);
  \draw[flow] (geom_c.east) -- (leak_c.west);

  \node[encbox, fit=(attn_c) (n2_c) (geom_c) (leak_c), label={[yshift=0.2cm]\footnotesize (c) GT-augmented (\texttt{geo\_causal=0})}] {};
\end{tikzpicture}

\vspace{0.6em}

\begin{tikzpicture}[node distance=0.62cm,scale=0.84,transform shape]
  \node[io] (in_d) {Decoder hidden state $h_t$};
  \node[hint, above=of in_d] (pred_d) {Predicted token dist.\\$p_t=\mathrm{softmax}(W_{\text{out}}h_t)$};
  \node[hint, right=2.5cm of pred_d] (next_d) {Mode E: \texttt{pred\_next\_detach}\\hint at $t \leftarrow p_tE$};
  \node[hint, above=of next_d] (prev_d) {Mode C: \texttt{pred\_prev\_causal\_detach}\\hint at $t \leftarrow p_{t-1}E$};
  \node[geom, above=of pred_d] (conv_d) {Geometric Conv Branch};
  \node[norm, above=of conv_d] (out_d) {Residual + Norm};

  \draw[flow] (in_d) -- (pred_d);
  \draw[flow] (pred_d) -- (conv_d);
  \draw[flow] (conv_d) -- (out_d);
  \draw[flow] (pred_d.east) -- (next_d.west);
  \draw[flow] (next_d.north) -- (prev_d.south);

  \node[encbox, fit=(in_d) (out_d) (next_d) (prev_d), label={[yshift=0.2cm]\footnotesize (d) Predicted-hint GT variants}] {};
\end{tikzpicture}

\caption{
\textbf{Geometric Transformer architectural variants studied in this paper.}}
\label{fig:gt-variant-family}
\end{figure*}

\paragraph{Categorical Interpretation of Auxiliary Channels}
\label{sec:gt-categorical-sidechannels}

The central issue is not whether attention is causal, but whether the \emph{full update diagram} at index $t$ includes extra morphisms that are future-informative. We formalize this at the diagram level. For a token sequence, let $X_i$ denote the representation object at position $i$ (after token and positional embedding, and any prefix-local processing up to that stage). Let $H_t$ denote the decoder hidden state object used to produce the logit at position $t$, and let $\hat y_t$ denote the output-logit object at position $t$. Thus, the base causal chain
\[
X_0 \to X_1 \to \cdots \to X_t \to H_t \to \hat y_t
\]
is an object-level shorthand for the prefix computation in the autoregressive model.

\paragraph{Base causal diagram.}
Let $\mathcal{I}_{\le t}$ be the prefix index category (objects are positions $0,\dots,t$ with arrows $i\!\to\!j$ for $i\le j$). A strict autoregressive decoder defines a functor
\[
F_t:\mathcal{I}_{\le t}\to \mathbf{Vect},
\]
and logits $\hat y_t$ are computed from the colimit object induced by this prefix-only diagram. In this setting, all computational paths factor through $\mathcal{I}_{\le t}$.

\paragraph{Auxiliary augmentation.}
Suppose we add an auxiliary object $A_t$ and morphism
\[
\alpha_t:A_t \to H_t,
\]
where $H_t$ is the decoder state at position $t$. This extends the diagram from a prefix-only shape to a larger category $\widetilde{\mathcal{I}}_t$.

\begin{definition}[Admissible Causal Augmentation]
\label{def:admissible-augmentation}
An auxiliary map $\alpha_t$ is \emph{admissible causal} iff it factors through the prefix diagram, i.e., there exists $G_t$ in the image of $F_t$ and maps
\[
A_t \to G_t \to H_t
\]
with $G_t$ constructed only from objects indexed by $\le t$.
\end{definition}

\begin{definition}[Future-Informative Augmentation]
\label{def:future-augmentation}
An auxiliary map $\alpha_t$ is \emph{future-informative} iff no such factorization exists through prefix-only objects, equivalently the induced update depends on objects indexed by $>t$ in $\widetilde{\mathcal{I}}_t$.
\end{definition}

Operationally, this means causal self-attention by itself is insufficient to guarantee strict causal evaluation: the geometric branch can alter the information regime depending on the source of its inputs.

\begin{figure*}[t]
\centering
\begin{subfigure}[t]{0.5\textwidth}
\centering
\[
\begin{tikzcd}[column sep=small, row sep=small]
X_0 \arrow[r] & X_1 \arrow[r] & \cdots \arrow[r] & X_t \arrow[r] & H_t \arrow[r] & \hat y_t
\end{tikzcd}
\]
\caption{Base causal diagram (prefix-only).}
\end{subfigure}
\hfill
\begin{subfigure}[t]{0.5\textwidth}
\centering
\[
\begin{tikzcd}[column sep=small, row sep=small]
X_0 \arrow[r] & \cdots \arrow[r] & X_t \arrow[r] & G_t \arrow[r, "\gamma_t"] & H_t \arrow[r] & \hat y_t \\
&& A_t \arrow[u, "\beta_t"] \arrow[urr, dashed, "\alpha_t"', bend left=18] &&
\end{tikzcd}
\]
\caption{Admissible causal augmentation: $\alpha_t=\gamma_t\!\circ\!\beta_t$.}
\end{subfigure}
\hfill
\begin{subfigure}[t]{0.5\textwidth}
\centering
\[
\begin{tikzcd}[column sep=small, row sep=small]
X_0 \arrow[r] & \cdots \arrow[r] & X_t \arrow[r] & H_t \arrow[r] & \hat y_t \\
X_{t+1} \arrow[r] & A_t \arrow[urr, "\alpha_t"', bend left=12] &&&
\end{tikzcd}
\]
\caption{Future-informative augmentation: $A_t$ depends on $X_{t+1}$.}
\end{subfigure}
\caption{\textbf{Information-regime diagrams for GT/Transformer updates.}
Left: strict causal prefix functor.
Middle: admissible auxiliary channel (Definition~\ref{def:admissible-augmentation}) factors through a prefix-derived object $G_t$, so the update remains in Regime C.
Right: future-informative auxiliary channel (Definition~\ref{def:future-augmentation}) keeps the same type $\alpha_t:A_t\!\to\!H_t$, but $A_t$ now depends on future-indexed objects, placing the model in Regime A.}
\label{fig:categorical-info-regimes}
\end{figure*}

\begin{table*}[b]
\centering
\small
\begin{tabular}{l l l}
\toprule
Variant & Auxiliary map type & Regime \\
\midrule
\texttt{transformer\_causal} & none beyond prefix functor & C \\
\texttt{gt\_causal} & causal local map (left-padded) & C \\
\texttt{gt\_pred\_prev\_causal\_detach} & shifted self-conditioning ($t\!\leftarrow\!t\!-\!1$) & C \\
\texttt{gt\_noncausal} & symmetric local map (can mix $t\!\pm\!1$) & A \\
\texttt{transformer\_future\_hint} & explicit future-informative hint path & A \\
\texttt{gt\_pred\_next\_detach} & same-index detached predicted-hint path & E \\
\bottomrule
\end{tabular}
\caption{\textbf{Model-to-regime mapping via auxiliary morphisms.}
The deciding factor is whether the auxiliary map factors through the prefix diagram.}
\label{tab:model-regime-mapping}
\end{table*}

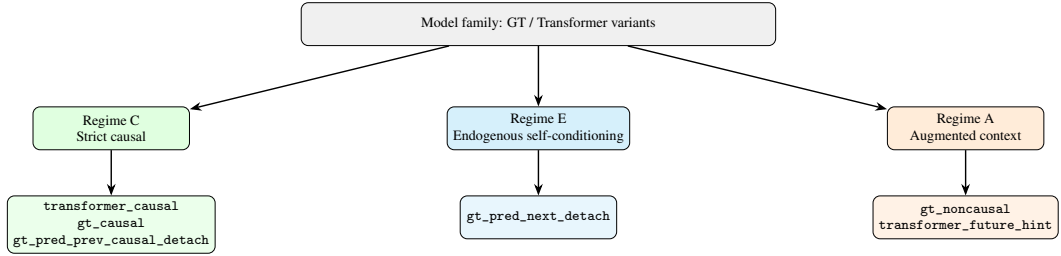
\begin{figure*}[t]
\centering
\resizebox{\linewidth}{!}{%
\begin{tikzpicture}[node distance=0.9cm]
  \node[block, fill=gray!12, minimum width=9.4cm] (root) {Model family: GT / Transformer variants};
  \node[block, fill=green!12, below left=1.2cm and 2.2cm of root, minimum width=3.1cm] (c) {Regime C\\Strict causal};
  \node[block, fill=cyan!14, below=1.2cm of root, minimum width=3.1cm] (e) {Regime E\\Endogenous self-conditioning};
  \node[block, fill=orange!15, below right=1.2cm and 2.2cm of root, minimum width=3.1cm] (a) {Regime A\\Augmented context};

  \node[block, fill=green!8, below=of c, minimum width=3.1cm] (c1) {\texttt{transformer\_causal}\\\texttt{gt\_causal}\\\texttt{gt\_pred\_prev\_causal\_detach}};
  \node[block, fill=cyan!8, below=of e, minimum width=3.1cm] (e1) {\texttt{gt\_pred\_next\_detach}};
  \node[block, fill=orange!10, below=of a, minimum width=3.1cm] (a1) {\texttt{gt\_noncausal}\\\texttt{transformer\_future\_hint}};

  \draw[flow] (root) -- (c);
  \draw[flow] (root) -- (e);
  \draw[flow] (root) -- (a);
  \draw[flow] (c) -- (c1);
  \draw[flow] (e) -- (e1);
  \draw[flow] (a) -- (a1);
\end{tikzpicture}
}
\caption{
\textbf{Evaluation taxonomy by information budget.}
Architecture comparisons are fair only within a regime; Regime E is reported separately from both strict-causal controls and explicit augmented-context leakage baselines.
}
\label{fig:gt-regime-taxonomy}
\end{figure*}

\subsection{Language-Model Instantiation}
\label{sec:gt-lm-instantiation}

For token sequence length $L$:
\begin{itemize}
  \item token positions $1,\dots,L$ are $0$-simplices;
  \item adjacency $(i,i+1)$ defines $1$-simplices;
  \item global operator is causal self-attention;
  \item geometric operator is a local 1D mixer (causal or symmetric depending on regime).
\end{itemize}

In the leakage-controlled variants we parameterize the geometric branch input source:
\begin{eqnarray*}
u_t \in
\left\{
h_t,\;
\mathrm{detach}(p_tE),\;
\mathrm{detach}(p_{t-1}E)
\right\}, \\ 
p_t=\mathrm{softmax}(W_{\text{out}}h_t/\tau).
\end{eqnarray*} 
The strict causal self-conditioned mode uses only $\mathrm{detach}(p_{t-1}E)$.

\paragraph{Which GT family is ablated here?}
All variants in this leakage protocol are GT-Lite style instantiations (Transformer + local depthwise Conv1D geometric mixer). They do not include GT-Full simplicial transport blocks or GT-MoE routing. This is intentional: the protocol isolates information-regime effects without adding capacity/routing confounds from the larger GT families.

\subsection{Revised LM Interpretation}

The revised empirical picture is:
\begin{itemize}
  \item In Regime C, \texttt{gt\_causal} and \texttt{transformer\_causal} are comparable; \texttt{gt\_pred\_prev\_causal\_detach} tracks this regime.
  \item In Regime E, \texttt{gt\_pred\_next\_detach} can move far below Regime C and, at larger width, approach the low-perplexity band of Regime A without becoming the same mechanism.
  \item In Regime A, \texttt{gt\_noncausal} and \texttt{transformer\_future\_hint} can show near-collapse in perplexity.
\end{itemize}
Therefore, large gains in Regime A are reported as augmented-information effects rather than architecture-only gains, while Regime E is treated separately as endogenous self-conditioning.

\paragraph{Reporting standard.}
Each LM table below includes an explicit regime label (C/E/A); the reproducibility materials record the corresponding \texttt{geo\_causal} flag, geometric input source, and shift policy for predicted hints.

\subsection{GT LM Results by Regime}

This section reports the PTB and WikiText-2 causal-ablation runs directly, and refers to the main paper for the corresponding WikiText-103 comparison, grouped by regime rather than by model name alone. The interpretation standard is:
\begin{enumerate}
\item make architecture claims only within the same information regime;
\item interpret cross-regime gains as information-budget effects.
\end{enumerate}

\paragraph{WikiText-2 (L=2, d=256) snapshot.}
The current WikiText-2 run shows the expected three-way split: strict-causal variants remain close to each other, the endogenous self-conditioning regime drops sharply below them, and the explicit augmented-context variants stay in the same low-perplexity band.
In this table, all \texttt{gt\_*} entries are GT-Lite style models (Transformer + local depthwise Conv1D geometric branch), not GT-Full or GT-MoE.

\begin{table}[h]
\centering
\small
\begin{tabular}{l c c c}
\toprule
Model & Regime & Val PPL & Test PPL \\
\midrule
\texttt{transformer\_causal} & C & 178.013 & 163.918 \\
\texttt{gt\_causal} & C & 169.707 & 157.744 \\
\texttt{gt\_pred\_prev\_causal\_detach} & C & 174.075 & 159.774 \\
\texttt{transformer\_future\_hint} & A & 2.073 & 2.023 \\
\texttt{gt\_pred\_next\_detach} & E & 1.624 & 1.593 \\
\texttt{gt\_noncausal} & A & 1.051 & 1.051 \\
\bottomrule
\end{tabular}
\caption{WikiText-2 causal-ablation results grouped by information regime.}
\label{tab:book-wikitext2-regime}
\end{table}

\paragraph{PTB (L=2, d=256) snapshot.}
PTB shows the same qualitative pattern: strict-causal variants cluster together, the endogenous self-conditioning regime nearly matches the leakage baselines, and the explicit augmented-context variants collapse perplexity.

\begin{table}[h]
\centering
\small
\begin{tabular}{l c c c}
\toprule
Model & Regime & Val PPL & Test PPL \\
\midrule
\texttt{transformer\_causal} & C & 134.767 & 124.474 \\
\texttt{gt\_causal} & C & 137.550 & 127.165 \\
\texttt{gt\_pred\_prev\_causal\_detach} & C & 134.735 & 126.499 \\
\texttt{transformer\_future\_hint} & A & 1.509 & 1.380 \\
\texttt{gt\_pred\_next\_detach} & E & 1.102 & 1.054 \\
\texttt{gt\_noncausal} & A & 1.075 & 1.050 \\
\bottomrule
\end{tabular}
\caption{PTB causal-ablation results grouped by information regime.}
\label{tab:book-ptb-regime}
\end{table}

\paragraph{The endogenous self-conditioning variant
(\texttt{gt\_pred\_next\_detach}).}
The most informative behavior in these tables is its transition profile: it
often begins close to strict-causal models and then shifts toward
augmented-context performance. The model uses detached predicted-token
embeddings to modulate a symmetric geometric branch. It does not inject a
future token embedding directly, but a carrier generated at a later source
position can still depend on tokens that are future information for an
earlier receiving target. We therefore interpret the result as endogenous
noncausal self-conditioning, not strict autoregressive performance.

\paragraph{Interpretive significance.}
Regime E remains mechanistically distinct from an explicit future embedding,
but it is also distinct from Regime C. The appropriate comparison therefore
separates (i) target-prefix-valid channels, (ii) endogenous but potentially
future-informative predictive channels, and (iii) explicitly augmented
channels. The second is a mechanism worth further study, but its perplexity
cannot be reported as strict-autoregressive evidence.

\paragraph{Smaller-model check (L=2, d=64).}
At lower width, \texttt{gt\_pred\_next\_detach} still improves strongly over Regime C baselines but no longer reaches Regime A performance. On PTB, test perplexity is $16.567$ for \texttt{gt\_pred\_next\_detach}, compared with $\sim 208$--$222$ (Regime C) and $\sim 1.144$--$4.940$ (Regime A). On WikiText-2, it reaches $44.050$, between $\sim 288$--$306$ (Regime C) and $\sim 1.421$--$10.160$ (Regime A). This indicates the transition behavior is likely capacity-dependent.

\begin{table}[h]
\centering
\small
\begin{tabular}{l l c c}
\toprule
Dataset & Model & Regime & Test PPL \\
\midrule
PTB ($d=64$) & \texttt{transformer\_causal} & C & 222.238 \\
PTB ($d=64$) & \texttt{gt\_causal} & C & 207.721 \\
PTB ($d=64$) & \texttt{gt\_pred\_prev\_causal\_detach} & C & 216.909 \\
PTB ($d=64$) & \texttt{gt\_pred\_next\_detach} & E & 16.567 \\
PTB ($d=64$) & \texttt{transformer\_future\_hint} & A & 4.940 \\
PTB ($d=64$) & \texttt{gt\_noncausal} & A & 1.144 \\
\midrule
WikiText-2 ($d=64$) & \texttt{transformer\_causal} & C & 305.511 \\
WikiText-2 ($d=64$) & \texttt{gt\_causal} & C & 288.030 \\
WikiText-2 ($d=64$) & \texttt{gt\_pred\_prev\_causal\_detach} & C & 295.857 \\
WikiText-2 ($d=64$) & \texttt{gt\_pred\_next\_detach} & E & 44.050 \\
WikiText-2 ($d=64$) & \texttt{transformer\_future\_hint} & A & 10.160 \\
WikiText-2 ($d=64$) & \texttt{gt\_noncausal} & A & 1.421 \\
\bottomrule
\end{tabular}
\caption{Small-model ($L=2$, $d=64$) causal-ablation snapshot.}
\label{tab:book-smallmodel-transition}
\end{table}

\paragraph{Capacity-transition index.}
Define
\[
\mathrm{transition\_gain}
=
\frac{\mathrm{PPL}_{\mathrm{causal\_best}}-\mathrm{PPL}_{\mathrm{pred\_next}}}
{\mathrm{PPL}_{\mathrm{causal\_best}}-\mathrm{PPL}_{\mathrm{aug\_best}}},
\]
where \(\mathrm{causal\_best}\) is the best strict-causal control and \(\mathrm{aug\_best}\) is the best augmented-context control. This measures how close \texttt{gt\_pred\_next\_detach} gets to augmented-context performance.

\begin{table}[h]
\centering
\small
\begin{tabular}{l c c c c}
\toprule
Dataset & \(d\) & Causal best & Pred-next & Transition gain \\
\midrule
PTB & 256 & 124.474 & 1.054 & 1.000 \\
WikiText-2 & 256 & 157.744 & 1.593 & 0.997 \\
PTB & 64 & 207.721 & 16.567 & 0.925 \\
WikiText-2 & 64 & 288.030 & 44.050 & 0.851 \\
\bottomrule
\end{tabular}
\caption{Capacity dependence of \texttt{gt\_pred\_next\_detach}: near-augmented performance at \(d=256\), but intermediate performance at \(d=64\).}
\label{tab:book-capacity-transition-gain}
\end{table}

\paragraph{WikiText-103.}
The corresponding WikiText-103 comparison is reported in the main paper, so we do not duplicate a second table here. The same qualitative pattern persists there: Regime C variants cluster together, Regime E remains substantially better than Regime C, and Regime A stays lowest because it admits future-informative side channels.

\paragraph{Interpretation and Scope}

The main empirical lesson is that the Geometric Transformer benefits should be interpreted as \emph{structure-aligned inductive bias}, not universal dominance. In regime-matched causal comparisons, gains are modest and task-dependent. In augmented-context settings, large gains can occur for multiple model families due to extra information pathways. 

\section{Algorithmic and Implementation Details} 

This appendix collects algorithmic and implementation details used in
the experiments.

\subsection{Quadratic Kan Extension Block}

\begin{algorithm}[h]
\caption{Quadratic Kan Extension Block}
\begin{algorithmic}[1]
\Require token states \(h_{0:S-1}\), simplex set \(\mathcal{I}\), value base \(v\), optional causal mask
\Ensure updated states \(h'_{0:S-1}\)
\State Construct simplex values \(V(\sigma)\) for all \(\sigma \in \mathcal{I}\)
\State Construct simplex keys \(K_\sigma = W_K V(\sigma)\)
\For{\(t = 0\) to \(S-1\)}
    \State \(Q_t \gets W_Q h_t\)
    \State \(\mathcal{I}_t \gets \{\sigma \in \mathcal{I} : \max(\sigma)\le t\}\) if causal mask is enabled;
           otherwise \(\mathcal{I}_t \gets \mathcal{I}\)
    \State \(w(t,\sigma) \gets \operatorname{softmax}_{\sigma \in \mathcal{I}_t}(Q_t^\top K_\sigma)\)
    \State \(m_t \gets \sum_{\sigma \in \mathcal{I}_t} w(t,\sigma)\,V(\sigma)\)
    \State \(h'_t \gets \mathrm{LN}\!\bigl(h_t + \mathrm{MLP}(m_t)\bigr)\)
\EndFor
\State \Return \(h'_{0:S-1}\)
\end{algorithmic}
\end{algorithm}

This block implements global simplicial pooling and scales
quadratically in sequence length under the present construction.

\subsection{Incidence-Restricted Kan Block}

\begin{algorithm}[h]
\caption{Incidence-Restricted Kan Block (Edge-Only)}
\begin{algorithmic}[1]
\Require token states \(h_{0:S-1}\), value base \(v_{0:S-1}\), regime flag
\Ensure updated states \(h'_{0:S-1}\)
\For{\(t = 1\) to \(S-1\)}
    \State \(e_t \gets \psi([v_{t-1},v_t])\)
\EndFor
\For{\(t = 0\) to \(S-1\)}
    \State \(m_t \gets 0\)
    \If{\(t \ge 1\)}
        \State \(m_t \gets m_t + \phi(e_t)\)
    \EndIf
    \If{regime is noncausal and \(t+1 \le S-1\)}
        \State \(m_t \gets m_t + \phi(e_{t+1})\)
    \EndIf
    \State \(h'_t \gets \mathrm{LN}(h_t + m_t)\)
\EndFor
\State \Return \(h'_{0:S-1}\)
\end{algorithmic}
\end{algorithm}

This block implements the linear-time approximation in which only
incident edges contribute to each token update.

\subsection{Predict--Detach Carrier Construction}

For self-conditioned regimes, the value base is not taken directly from
teacher-forced hidden states at future positions.
Instead we construct detached predictive carriers:
\begin{equation}
\hat e_t
=
\operatorname{detach}\!\Bigl(
\operatorname{softmax}(\ell_t/T)\,E
\Bigr),
\qquad
\ell_t = W_o h_t.
\label{eq:appendix_pred_detach}
\end{equation}

These carriers may then be used in either quadratic or
incidence-restricted Kan aggregation, as well as in TopoCoend
neighborhood construction.

\subsection{Leakage Test Procedure}

To verify that a regime does not exploit gold future-token leakage, we
use the following diagnostic.

\begin{enumerate}
    \item Randomly shuffle the target tokens \(x_{t+1}\) within a batch.
    \item Train the model for a short schedule under the candidate regime.
    \item Observe whether perplexity remains sensible or collapses.
\end{enumerate}

In explicitly gold-noncausal regimes, perplexity collapses toward \(1\)
even under shuffled targets, indicating that the model is exploiting an
invalid information path.
In predict-and-detach regimes, perplexity remains high under shuffling,
supporting the claim that no direct gold-future leakage is present.

\subsection{Hyperparameters and Training Details}

The released experiments fall into two families.

\paragraph{Main causal language-model comparisons.}
The summary tables and plots under \texttt{plots/PTB\_Spark},
\texttt{plots/Wiki-2-Spark}, and \texttt{plots/Wiki103\_Spark} are
produced by \texttt{scripts/run\_lm\_causal\_comparison\_kan\_topocoend.py}
and the corresponding training scripts it invokes. These runs use:
\begin{itemize}
    \item optimizer: AdamW
    \item learning rate: \(3\times 10^{-4}\)
    \item weight decay: \(10^{-5}\)
    \item batch size: \(32\)
    \item context length: \(128\)
    \item number of layers: \(2\)
    \item hidden dimension: \(256\)
    \item attention heads: \(4\)
    \item training iterations: \(5000\)
    \item reporting cadence: every \(100\) steps
    \item evaluation cadence: every \(1000\) steps
\end{itemize}

Self-conditioned regimes use temperature \(T=1.0\) unless otherwise
noted. TopoCoend uses \(\texttt{topo\_k}=16\) and
\(\texttt{topo\_dim}=16\). The released LM figures are single-seed runs,
consistent with the checklist statement that we do not report
multi-seed error bars.

\paragraph{Structured completion runs.}
The block-completion results are produced by
\texttt{scripts/ket\_experiments/ket\_unified\_harness.py} and
\texttt{scripts/ket\_experiments/train/train\_ket\_block.py}. The
reported PTB/WT2/WT103 higher-depth block comparisons use:
\begin{itemize}
    \item matched Transformer and incidence-KET backbones
    \item optimizer: AdamW
    \item learning rate: \(3\times 10^{-4}\)
    \item weight decay: \(10^{-2}\)
    \item context length: \(128\)
    \item hidden dimension: \(64\)
    \item layers: \(8\) and \(16\)
    \item attention heads: \(4\)
    \item block size: \(4\)
    \item training steps: \(2000\)
    \item evaluation cadence: every \(250\) steps
    \item denoising steps: \(8\)
    \item evaluation cap: \(100\) held-out batches
\end{itemize}

These structured-completion runs use the local benchmark text files under
\texttt{scripts/ket\_experiments/data/} and apply the same GPT-2
tokenizer across the matched backbones.

Gradients are clipped to norm \(1.0\).
The released block runs record seeds \(7, 11, 17, 19,\) and \(1337\) in the accompanying
\texttt{all\_runs.csv} files.

\subsection{Hardware and Runtime Measurement}

The figures and tables in the main LM comparison were generated from the
released DGX Spark runs in \texttt{plots/*\_Spark}. Additional pilot runs
were also exercised on Apple MPS hardware during development. The
structured-completion harness records the same runtime fields in
\texttt{scripts/ket\_experiments/results/logs/}. The hardware families
used in the released artifacts are:
\begin{itemize}
    \item Apple MPS hardware
    \item NVIDIA DGX Spark (Blackwell GPU)
\end{itemize}

Runtime is measured end-to-end by wall-clock time.
Iterations per second are computed as
\[
\text{iters/sec}
=
\frac{\text{number of iterations}}{\text{total wall-clock seconds}}.
\]

Quadratic KET benefits disproportionately from modern GPU acceleration,
which significantly narrows its runtime gap relative to incidence-based
or baseline models.

\section{Broader Impact} 
\label{app:broader_impact} 

This paper describes a new family of Kan Extension Transformers, which are useful in building structured language models. The applicability of this framework extends to all of language models, and consequently carries the usual societal risk of deployment of models that might hallucinate or behave in ways that are undesirable. The focus of this paper was building the basic methodology, and not on fine-tuning the model to obtain satisfactory performance in particular regimes.

\end{document}